\documentclass{llncs}

\usepackage[linesnumbered,lined,boxed,vlined]{algorithm2e}
\usepackage{amsfonts}
\usepackage{amsmath}
\usepackage{arydshln}  % dash line in table
\usepackage[table]{xcolor}  % for color in table
\usepackage{graphics}  % for scalebox
\usepackage{multirow}  % for multirow in table 
\def\b{\ensuremath\boldsymbol}
\usepackage{graphicx}

\usepackage{tikz}
\usepackage{lipsum}
\newcommand\copyrighttextt{%
  \footnotesize Published at International Conference on Image Analysis and Recognition, Springer. This version includes the supplementary material for derivation of some equations.}
\newcommand\copyrightnotice{%
\begin{tikzpicture}[remember picture,overlay]
\node[anchor=south,yshift=10pt] at (current page.south) {\fbox{\parbox{\dimexpr\textwidth-\fboxsep-\fboxrule\relax}{\copyrighttextt}}};
\end{tikzpicture}%
}

\begin{document}

\title{Locally Linear Image Structural Embedding \\for Image Structure Manifold Learning}
\author{Benyamin Ghojogh, 
Fakhri Karray,
Mark Crowley
}

\institute{Department of Electrical and Computer Engineering,\\ University of Waterloo, Waterloo, ON, Canada  \\
\email{\{bghojogh, karray, mcrowley\}@uwaterloo.ca}
}

\maketitle              % typeset the title of the contribution

\begin{abstract}
Most of existing manifold learning methods rely on Mean Squared Error (MSE) or $\ell_2$ norm. However, for the problem of image quality assessment, these are not promising measure. In this paper, we introduce the concept of an image structure manifold which captures image structure features and discriminates image distortions. We propose a new manifold learning method, Locally Linear Image Structural Embedding (LLISE), and kernel LLISE for learning this manifold. The LLISE is inspired by Locally Linear Embedding (LLE) but uses SSIM rather than MSE. This paper builds a bridge between manifold learning and image fidelity assessment and it can open a new area for future investigations.
\end{abstract}

\copyrightnotice

\section{Introduction}

Mean Squared Error (MSE) is not a good measure for image quality assessment \cite{wang2009mean}. Two different categories of distortions exist, i.e., structural and non-structural distortions \cite{wang2004image}.
The structural similarity index (SSIM) \cite{wang2004image,wang2006modern} is found to be a very promising measure for image fidelity assessment. It encounters luminance and contrast change as non-structural distortions and other distortions as structural ones. Recently, it has been used in optimization problems for different tasks although it is not convex but quasi-convex under certain conditions \cite{brunet2012mathematical}.

The manifold learning methods are designed mostly based on MSE or the $\ell_2$ norm. Therefore, they do not perform satisfactorily for image quality discrimination. Locally Linear Embedding (LLE) \cite{roweis2000nonlinear} is an example. In this paper, we introduce the new concept of \textit{image structure manifold} which captures the features of image structure and is useful for discriminating the image distortions. 
We propose Locally Linear Image Structural Embedding (LLISE), in both original and feature space, which uses SSIM distance rather than $\ell_2$ norm. We also propose the out-of-sample extension of LLISE.
The derivations of expressions in this paper are detailed more in the supplementary-material paper which will be released in \url{https://arXiv.org}.

\subsection{Structural Similarity Index}

The SSIM between two reshaped image blocks $\breve{\b{x}}_1 = [x_1^{[1]}, \dots, x_1^{[q]}]^\top \in \mathbb{R}^q$ and $\breve{\b{x}}_2 = [x_2^{[1]}, \dots, x_2^{[q]}]^\top \in \mathbb{R}^q$, in color intensity range $[0,l]$, is \cite{wang2004image,wang2006modern}:
\begin{align}
\mathbb{R} \ni \text{SSIM}(\breve{\b{x}}_1, \breve{\b{x}}_2) := \bigg(\frac{2 \mu_{x_1} \mu_{x_2} + c_1}{\mu_{x_1}^2 + \mu_{x_2}^2 + c_1}\bigg) \bigg(\frac{2 \sigma_{x_1} \sigma_{x_2} + c_2}{\sigma_{x_1}^2 + \sigma_{x_2}^2 + c_2}\bigg) \bigg(\frac{\sigma_{x_1, x_2} + c_3}{\sigma_{x_1}\sigma_{x_2} + c_3}\bigg),
\end{align}
where $\mu_{x_1} = (1/q) \sum_{i=1}^q x_1^{[i]}$, $\sigma_{x_1} = \Big[\big(1/(q-1)\big) \sum_{i=1}^q (x_1^{[i]} - \mu_{x_1})^2\Big]^{0.5}$, $\sigma_{x_1,x_2} = \big(1/(q-1)\big) \sum_{i=1}^q (x_1^{[i]} - \mu_{x_1}) (x_2^{[i]} - \mu_{x_2})$, $c_1=(0.01 \times l)^2$, $c_2=2\,c_3=(0.03 \times l)^2$, and $\mu_{x_2}$ and $\sigma_{x_2}$ are defined similarly for $\breve{\b{x}}_2$. In this work, $l=1$.

Because of $c_2=2\,c_3$, the SSIM is simplified to $\text{SSIM}(\breve{\b{x}}_1, \breve{\b{x}}_2) = s_1(\breve{\b{x}}_1, \breve{\b{x}}_2) \times s_2(\breve{\b{x}}_1, \breve{\b{x}}_2)$, where $s_1(\breve{\b{x}}_1, \breve{\b{x}}_2) := (2 \mu_{x_1} \mu_{x_2} + c_1)/(\mu_{x_1}^2 + \mu_{x_2}^2 + c_1)$ and $s_2(\breve{\b{x}}_1, \breve{\b{x}}_2) := (2 \sigma_{x_1, x_2} + c_2)/(\sigma_{x_1}^2 + \sigma_{x_2}^2 + c_2)$.
If the vectors $\breve{\b{x}}_1$ and $\breve{\b{x}}_2$ have zero mean, i.e., $\mu_{x_1} = \mu_{x_2} = 0$, the SSIM becomes $\mathbb{R} \ni \text{SSIM}(\breve{\b{x}}_1, \breve{\b{x}}_2) = (2\breve{\b{x}}_1^\top \breve{\b{x}}_2 + c) / (||\breve{\b{x}}_1||_2^2 + ||\breve{\b{x}}_2||_2^2 + c)$, where $c = (q-1) \,c_2$ \cite{otero2014unconstrained}. 
We denote the reshaped vectors of the two images by $\b{x}_1 \in \mathbb{R}^{d}$ and $\b{x}_2 \in \mathbb{R}^{d}$, and a reshaped block in the two images by $\breve{\b{x}}_1 \in \mathbb{R}^{q}$ and $\breve{\b{x}}_2 \in \mathbb{R}^{q}$.
If $\mu_{x_1} = \mu_{x_2} = 0$, the (squared) distance based on SSIM, which we denote by $||.||_S$, is \cite{brunet2012mathematical,otero2014unconstrained}:
\begin{align}\label{equation_SSIM_distance}
\mathbb{R} \ni ||\breve{\b{x}}_1 - \breve{\b{x}}_2||_S := 1 -  \text{SSIM}(\breve{\b{x}}_1, \breve{\b{x}}_2) = \frac{||\breve{\b{x}}_1 - \breve{\b{x}}_2||_2^2}{||\breve{\b{x}}_1||_2^2 + ||\breve{\b{x}}_2||_2^2 + c}.
\end{align}

\subsection{Locally Linear Embedding}

In LLE \cite{roweis2000nonlinear}, first a $k$-Nearest Neighbor ($k$-NN) graph is found using pairwise Euclidean distances. Every data point $\b{x}_j \in \mathbb{R}^d$ is reconstructed by its $k$ neighbors $\mathbb{R}^{d \times k} \ni \b{X}_{j} := [ \,_1\b{x}_{j}, \dots,\, _k\b{x}_{j}]$ where $_r\b{x}_{j}$ denotes the $r$-th neighbor of $\b{x}_j$. If $\mathbb{R}^k \ni \widetilde{\b{w}}_{j} := [\,_1\widetilde{w}_{j}, \dots, \,_k\widetilde{w}_{j}]^\top$ denotes the reconstruction weights for the $\b{x}_j$, the reconstruction problem with the weights adding to one is: minimize $\sum_{j=1}^n ||\b{x}_j - \sum_{r=1}^k \,_r\widetilde{w}_{j} \,_r\b{x}_{j}||_2^2$, subject to $\sum_{r=1}^k \,_r\widetilde{w}_{j} = 1, \forall j \in \{1, \dots, n\}$.
Then, the data points are embedded using the obtained weights. Take $\mathbb{R}^n \ni \b{w}_{j} := [\,_1w_{j}, \dots, \,_nw_{j}]^\top$ where $_rw_{j}$ is the weight obtained from linear reconstruction if $\b{x}_r$ is a neighbor of $\b{x}_j$ and is zero otherwise. If $\b{y}_j \in \mathbb{R}^p$ denotes the embedded $j$-th data point, the embedding problem with unit covariance is: minimize $\sum_{j=1}^n ||\b{y}_j - \sum_{r=1}^n \,_rw_{j}\, \b{y}_r||_2^2$, subject to $(1/n) \sum_{j=1}^n \b{y}_j \b{y}_j^\top = \b{I}$ and $\sum_{j=1}^n \b{y}_j = \b{0}, \forall j \in \{1, \dots, n\}$.
Kernel LLE \cite{zhao2012facial} finds the $k$-NN graph and performs linear reconstruction from the neighbors in the feature space.

\section{Locally Linear Image Structural Embedding}

We partition a $d$-dimensional image $\b{x}$ into $b = \lceil d / q \rceil$ non-overlapping blocks each of which is a reshaped vector $\breve{\b{x}} \in \mathbb{R}^q$. 
In LLISE, we find a $p$-dimensional image structure manifold for every block.
The $q$ is a parameter and is an upper bound on the desired dimensionality of the manifold of a block ($p \leq q$). This parameter is better not to be a very large number because of spatial variety of image statistics, and not very small to be able to capture the image structure. 
We denote the $i$-th block in the $j$-th image by $\breve{\b{x}}_{j,i} \in \mathbb{R}^q$.
In LLISE, we first center every image block by removing its mean.

\subsection{Embedding The Training Data}

\subsubsection{$k$-Nearest Neighbors}

For every block $\breve{\b{x}}_i$ ($i \in \{1, \dots,b\}$), amongst the $n$ images, a $k$-NN graph is formed using pairwise Euclidean distances between that $i$-th block in the $n$ images. Therefore, every block in every image has $k$ neighbors. Let $_r\breve{\b{x}}_{j,i} \in \mathbb{R}^q$ denote the $r$-th neighbor of $\breve{\b{x}}_{j,i}$ and let the matrix $\mathbb{R}^{q \times k} \ni \breve{\b{X}}_{j,i} := [ \,_1\breve{\b{x}}_{j,i}, \dots,\, _k\breve{\b{x}}_{j,i}]$ include the neighbors of $\breve{\b{x}}_{j,i}$.

% We also take $\mathbb{R}^{q \times n} \ni \breve{\b{X}}_i := [\breve{\b{x}}_{1,i}, \dots, \breve{\b{x}}_{n,i}]$ to denote the matrix including the $i$-th block in images.

\subsubsection{Linear Reconstruction by the Neighbors}

For every block $\breve{\b{x}}_i$, we want the $j$-th image to be linearly reconstructed by its $k$ neighbors. We minimize the reconstruction error while the vector of reconstruction weights for every image block is a unit vector:
\begin{equation}\label{equation_LLISE_linearReconstruct}
\begin{aligned}
& \underset{\widetilde{\b{W}}_i}{\text{minimize}}
& & \sum_{i=1}^b \varepsilon(\widetilde{\b{W}}_i) := \sum_{i=1}^b \sum_{j=1}^n \Big|\Big|\, \breve{\b{x}}_{j,i} - \sum_{r=1}^k \,_r\widetilde{w}_{j,i}\, \,_r\breve{\b{x}}_{j,i}\Big|\Big|_S, \\
& \text{subject to}
& & \sum_{r=1}^k \,_r\widetilde{w}_{j,i}^2 = 1, ~~~ \forall i \in \{1, \dots, b\}, ~~ \forall j \in \{1, \dots, n\},
\end{aligned}
\end{equation}
where $\mathbb{R}^{n \times k} \ni \widetilde{\b{W}}_i := [\widetilde{\b{w}}_{1,i}, \dots, \widetilde{\b{w}}_{n,i}]^\top$ includes the weights for the $i$-th block in the images and $\mathbb{R}^k \ni \widetilde{\b{w}}_{j,i} := [\,_1\widetilde{w}_{j,i}, \dots, \,_k\widetilde{w}_{j,i}]^\top$ includes the weights of linear reconstruction of the $i$-th block in the $j$-th image using its $k$ neighbors.
The constraint ensures $\widetilde{\b{w}}_{j,i}^\top \widetilde{\b{w}}_{j,i} = ||\widetilde{\b{w}}_{j,i}||_2^2 = 1$.
Note that we can formulate the problem with the constraint $\sum_{r=1}^k \,_r\widetilde{w}_{j,i} = 1$ as in LLE; however, with that constraint, the weights start to explode gradually after some optimization iterations. This problem does not happen in LLE because LLE is solved in closed form and not iteratively.

Take $f(\widetilde{\b{w}}_{j,i}) := \big|\big|\breve{\b{x}}_{j,i} - \sum_{r=1}^k \,_r\widetilde{w}_{j,i}\, _r\breve{\b{x}}_{j,i}\big|\big|_S$ which is restated as $f(\widetilde{\b{w}}_{j,i}) = ||\breve{\b{x}}_{j,i} - \breve{\b{X}}_{j,i}\, \widetilde{\b{w}}_{j,i}||_S$.
According to Eq. (\ref{equation_SSIM_distance}), the $f(\widetilde{\b{w}}_{j,i})$ is simplified to:
\begin{align}\label{equation_LLISE_f}
\mathbb{R} \ni f(\widetilde{\b{w}}_{j,i}) = \frac{\breve{\b{x}}_{j,i}^\top\, \breve{\b{x}}_{j,i} + \widetilde{\b{w}}_{j,i}^\top\, \breve{\b{X}}_{j,i}^\top\, \breve{\b{X}}_{j,i}\, \widetilde{\b{w}}_{j,i} - 2\, \widetilde{\b{w}}_{j,i}^\top\, \breve{\b{X}}_{j,i}^\top\, \breve{\b{x}}_{j,i}}{\breve{\b{x}}_{j,i}^\top\, \breve{\b{x}}_{j,i} + \widetilde{\b{w}}_{j,i}^\top\, \breve{\b{X}}_{j,i}^\top\, \breve{\b{X}}_{j,i}\, \widetilde{\b{w}}_{j,i} + c}.
\end{align}
The gradient of $f(\widetilde{\b{w}}_{j,i})$ with respect to $\widetilde{\b{w}}_{j,i}$ is:
\begin{align}\label{equation_LLISE_derivative_f}
\mathbb{R}^k \ni \nabla f(\widetilde{\b{w}}_{j,i}) = \frac{2\, \breve{\b{X}}_{j,i}^\top \Big(\big(1 - f(\widetilde{\b{w}}_{j,i})\big) \breve{\b{X}}_{j,i} \widetilde{\b{w}}_{j,i} - \breve{\b{x}}_{j,i}\Big)}{\breve{\b{x}}_{j,i}^\top\, \breve{\b{x}}_{j,i} + \widetilde{\b{w}}_{j,i}^\top\, \breve{\b{X}}_{j,i}^\top\, \breve{\b{X}}_{j,i}\, \widetilde{\b{w}}_{j,i} + c}.
\end{align}
The Eq. (\ref{equation_LLISE_linearReconstruct}) can be rewritten as:
\begin{equation}\label{equation_LLISE_linearReconstruct_2}
\begin{aligned}
& \underset{\widetilde{\b{w}}_{j,i},\, \widetilde{\b{\xi}}_{j,i}}{\text{minimize}}
& & \sum_{i=1}^b \sum_{j=1}^n \big( f(\widetilde{\b{w}}_{j,i}) + h_1(\widetilde{\b{\xi}}_{j,i}) \big), \\
& \text{subject to}
& & \widetilde{\b{w}}_{j,i} - \widetilde{\b{\xi}}_{j,i} = 0 ~~ \forall i \in \{1, \dots, b\}, ~~ \forall j \in \{1, \dots, n\},
\end{aligned}
\end{equation}
where $\mathbb{R}^k \ni \widetilde{\b{\xi}}_{j,i} := [_1\widetilde{\xi}_{j,i}, \dots, \,_k\widetilde{\xi}_{j,i}]^\top$ and $h_1(\widetilde{\b{\xi}}_{j,i}) := \mathbb{I}\big(\widetilde{\b{\xi}}_{j,i}^\top\, \widetilde{\b{\xi}}_{j,i} = 1\big)$. The $\mathbb{I}(.)$ denotes the indicator function which is zero if its condition is satisfied and is infinite otherwise. 
The Eq. (\ref{equation_LLISE_linearReconstruct_2}) can be solved using Alternating Direction Method of Multipliers (ADMM) \cite{boyd2011distributed,otero2018alternate}.
The augmented Lagrangian is: $\mathcal{L}_{\rho} = \sum_{i=1}^b \sum_{j=1}^n \big( f(\widetilde{\b{w}}_{j,i}) + h_1(\widetilde{\b{\xi}}_{j,i}) \big) + (\rho/2)\, ||\widetilde{\b{w}}_{j,i} - \widetilde{\b{\xi}}_{j,i} + \b{j}_{j,i}||_2^2 - (\rho/2)\, ||\b{\lambda}_{j,i}||_2^2$, where $\b{\lambda}_{j,i} \in \mathbb{R}^k$ is the Lagrange multiplier, $\rho > 0$ is a parameter, and $\mathbb{R}^k \ni \b{j}_{j,i} := (1/\rho) \b{\lambda}_{j,i}$. The term $(\rho/2)\, ||\b{\lambda}_{j,i}||_2^2$ is a constant with respect to $\widetilde{\b{w}}_{j,i}$ and $\widetilde{\b{\xi}}_{j,i}$ and can be dropped. 
The updates of $\widetilde{\b{w}}_{j,i}$, $\widetilde{\b{\xi}}_{j,i}$, and $\b{j}_{j,i}$ are performed as \cite{boyd2011distributed,otero2018alternate}:
\begin{align}
\widetilde{\b{w}}_{j,i}^{(\nu+1)} & := \arg \min_{\widetilde{\b{w}}_{j,i}} \Big( f(\widetilde{\b{w}}_{j,i}) + (\rho/2)\, ||\widetilde{\b{w}}_{j,i} - \widetilde{\b{\xi}}_{j,i}^{(\nu)} + \b{j}_{j,i}^{(\nu)}||_2^2 \Big), \label{equation_LLISE_reconst_ADMM_w_update} \\
\widetilde{\b{\xi}}_{j,i}^{(\nu+1)} & := \arg \min_{\widetilde{\b{\xi}}_{j,i}} \Big( h_1(\widetilde{\b{\xi}}_{j,i}) + (\rho/2)\, ||\widetilde{\b{w}}_{j,i}^{(\nu+1)} - \widetilde{\b{\xi}}_{j,i} + \b{j}_{j,i}^{(\nu)}||_2^2 \Big), \label{equation_LLISE_reconst_ADMM_xi_update} \\
\b{j}_{j,i}^{(\nu+1)} & := \b{j}_{j,i}^{(\nu)} + \widetilde{\b{w}}_{j,i}^{(\nu+1)} - \widetilde{\b{\xi}}_{j,i}^{(\nu+1)}, \label{equation_LLISE_reconst_ADMM_j_update}
\end{align}
where $\nu$ denotes the iteration.
The gradient of the objective function in Eq. (\ref{equation_LLISE_reconst_ADMM_w_update}) is $\nabla f(\widetilde{\b{w}}_{j,i}) + \rho\, (\widetilde{\b{w}}_{j,i} - \widetilde{\b{\xi}}_{j,i}^{(\nu)} + \b{j}_{j,i}^{(\nu)})$. We can use the gradient decent method \cite{boyd2004convex} for solving the Eq. (\ref{equation_LLISE_reconst_ADMM_w_update}). Our experiments showed that even one iteration of gradient decent suffices for Eq. (\ref{equation_LLISE_reconst_ADMM_w_update}) because the ADMM itself is iterative. Hence, we can replace this equation with one iteration of gradient decent. 

The proximal operator is defined as \cite{parikh2014proximal}:
\begin{align}\label{equation_prox_vector}
\textbf{prox}_{\lambda, h}(\b{v}) := \arg \min_{\b{u}}\,\big(h(\b{u}) + (\lambda/2) ||\b{u} - \b{v}||_2^2\big),
\end{align}
where $\lambda$ is the proximal parameter and $h$ is the function that the proximal algorithm wants to minimize.
According to Eq. (\ref{equation_prox_vector}), the Eq. (\ref{equation_LLISE_reconst_ADMM_xi_update}) is equivalent to $\textbf{prox}_{\rho, h_1}(\widetilde{\b{w}}_{j,i}^{(\nu+1)} + \b{j}_{j,i}^{(\nu)})$. 
As $h_1(.)$ is indicator function, its proximal operator is projection \cite{parikh2014proximal}. Therefore, Eq. (\ref{equation_LLISE_reconst_ADMM_xi_update}) is equivalent to $\mathrm{\Pi}(\widetilde{\b{w}}_{j,i}^{(\nu+1)} + \b{j}_{j,i}^{(\nu)})$ where $\mathrm{\Pi}(.)$ denotes projection onto a set. 
The condition in $h_1(.)$ is $\widetilde{\b{\xi}}_{j,i}^\top \widetilde{\b{\xi}}_{j,i} = 1$; therefore, this projection normalizes the vector by dividing to its $\ell_2$ norm.

In summary, the Eqs. (\ref{equation_LLISE_reconst_ADMM_w_update}), (\ref{equation_LLISE_reconst_ADMM_xi_update}), and (\ref{equation_LLISE_reconst_ADMM_j_update}) can be restated as:
\begin{equation}\label{equation_LLISE_reconst_ADMM_updates}
\begin{aligned}
\widetilde{\b{w}}_{j,i}^{(\nu+1)} & := \widetilde{\b{w}}_{j,i}^{(\nu)} - \eta\, \nabla f(\widetilde{\b{w}}_{j,i}^{(\nu)}) - \eta\,\rho\, (\widetilde{\b{w}}_{j,i}^{(\nu)} - \widetilde{\b{\xi}}_{j,i}^{(\nu)} + \b{j}_{j,i}^{(\nu)}), \\
\widetilde{\b{\xi}}_{j,i}^{(\nu+1)} & := (\widetilde{\b{w}}_{j,i}^{(\nu+1)} + \b{j}_{j,i}^{(\nu)}) / ||\widetilde{\b{w}}_{j,i}^{(\nu+1)} + \b{j}_{j,i}^{(\nu)}||_2 , \\
\b{j}_{j,i}^{(\nu+1)} & := \b{j}_{j,i}^{(\nu)} + \widetilde{\b{w}}_{j,i}^{(\nu+1)} - \widetilde{\b{\xi}}_{j,i}^{(\nu+1)},
\end{aligned}
\end{equation}
where $\eta > 0$ is the learning rate.
Iteratively solving Eq. (\ref{equation_LLISE_reconst_ADMM_updates}) until convergence gives us the $\widetilde{\b{w}}_{j,i}$ for the $i$-th block in the $j$-th image. 
Note that Eq. (\ref{equation_LLISE_reconst_ADMM_updates}) can be solved in parallel for the blocks of images.

\subsubsection{Linear Embedding}

In the previous section, we found the weights of linear reconstruction of the $i$-th block in every image from the $i$-th block in its $k$-NN. We can now find the embedding of the $i$-th block in every image using the obtained weights of reconstruction:
\begin{equation}\label{equation_LLISE_linearEmbedding}
\begin{aligned}
& \underset{\b{Y}_i}{\text{minimize}}
& & \sum_{i=1}^b \sum_{j=1}^n \big|\big|\b{y}_{j,i} - \sum_{r=1}^n \,_rw_{j,i}\, \b{y}_{r,i}\big|\big|_S, \\
& \text{subject to}
& & \frac{1}{n} \sum_{j=1}^n \b{y}_{j,i} \b{y}_{j,i}^\top = \b{I}, ~~~ \sum_{j=1}^n \b{y}_{j,i} = \b{0}, ~~~ \forall i \in \{1, \dots, b\}, 
\end{aligned}
\end{equation}
where $\b{I}$ is the identity matrix, the rows of $\mathbb{R}^{n \times p} \ni \b{Y}_i := [\b{y}_{1,i}, \dots, \b{y}_{n,i}]^\top$ are the embedded $i$-th block in the images, $\b{y}_{r,i} \in \mathbb{R}^p$ is the $i$-th embedded block in the $r$-th image, and $_rw_{j,i}$ is the weight obtained from the linear reconstruction (previous section) if $\b{x}_{r,i}$ is a neighbor of $\b{x}_{j,i}$ and zero otherwise.
The second constraint ensures the zero mean of embedded blocks. The first and second constraints together satisfy having unit covariance for the embedded image blocks.

Suppose $\mathbb{R}^n \ni \b{w}_{j,i} := [\,_1w_{j,i}, \dots, \,_nw_{j,i}]^\top$ and let $\mathbb{R}^n \ni \b{1}_j := [0, \dots, 1, \dots, 0]^\top$ be the vector whose $j$-th element is one and other elements are zero.
The Eq. (\ref{equation_LLISE_linearEmbedding}) can be restated as:
\begin{equation}\label{equation_LLISE_linearEmbedding_2}
\begin{aligned}
& \underset{\b{Y}_i}{\text{minimize}}
& & \sum_{i=1}^b \sum_{j=1}^n ||\b{Y}_i^\top \b{1}_j - \b{Y}_i^\top \b{w}_{j,i}||_S, \\
& \text{subject to}
& & \frac{1}{n} \b{Y}_i^\top \b{Y}_i = \b{I}, ~~~ \b{Y}_i^\top \b{1} = \b{0}, ~~~ \forall i \in \{1, \dots, b\}.
\end{aligned}
\end{equation}
Let $\theta_j(\b{\b{Y}}_i) := ||\b{Y}_i^\top \b{1}_j - \b{Y}_i^\top \b{w}_{j,i}||_S$. According to Eq. (\ref{equation_SSIM_distance}), it is simplified to:
\begin{align}\label{equation_LLISE_theta_Y}
\mathbb{R} \ni \theta_j(\b{Y}_i) = \frac{\textbf{tr}(\b{Y}_i^\top \b{M}_{j,i}\, \b{Y}_i)}{\textbf{tr}(\b{Y}_i^\top \b{\Psi}_{j,i}\, \b{Y}_i) + c},
\end{align}
where $\textbf{tr}(.)$ is the trace of matrix, $\mathbb{R}^{n \times n} \ni \b{M}_{j,i} := \b{1}_j \b{1}_j^\top + \b{w}_{j,i} \b{w}_{j,i}^\top - 2\, \b{1}_j \b{w}_{j,i}^\top$, and $\mathbb{R}^{n \times n} \ni \b{\Psi}_{j,i} := \b{1}_j \b{1}_j^\top + \b{w}_{j,i} \b{w}_{j,i}^\top = \b{M}_{j,i} + 2\, \b{1}_j \b{w}_{j,i}^\top$.
The gradient of $\theta_j(\b{Y}_i)$ with respect to $\b{Y}_i$ is:
\begin{align}\label{equation_LLISE_gradient_theta_Y}
\mathbb{R}^{n \times p} \ni \nabla \theta_j(\b{Y}_i) = \frac{2}{\textbf{tr}(\b{Y}_i^\top \b{\Psi}_{j,i}\, \b{Y}_i) + c} \Big(\b{M}_{j,i} - \theta_j(\b{Y}_i)\,\b{\Psi}_{j,i}\Big)\, \b{Y}_i.
\end{align}

In Eq. (\ref{equation_LLISE_linearEmbedding_2}), we can embed the constraint as an indicator function in the objective function \cite{boyd2011distributed}:
\begin{equation}\label{equation_LLISE_linearEmbedding_3}
\begin{aligned}
& \underset{\b{Y}_i, \b{V}_i \in \mathbb{R}^{n \times p}}{\text{minimize}}
& & \sum_{i=1}^b \Big( \sum_{j=1}^n \big( \theta_j(\b{Y}_i) \big) + h_2(\b{V}_i) \Big), \\
& \text{subject to}
& & \b{Y} - \b{V} = \b{0}, 
\end{aligned}
\end{equation}
where $h_2(\b{V}_i) := \mathbb{I}\big(\b{V}_i^\top \b{1} = \b{0} \wedge (1/n)\b{V}_i^\top \b{V}_i = \b{I}\big)$. The $\b{U}$ and $\b{V}$ are union of partitions, i.e., $\b{Y} := \cup_{i=1}^b \b{Y}_i$ and $\b{V} := \cup_{i=1}^b \b{V}_i$ \cite{otero2018alternate}.

We can solve the Eq. (\ref{equation_LLISE_linearEmbedding_3}) using Alternating Direction Method of Multipliers (ADMM) \cite{boyd2011distributed,otero2018alternate}.
The augmented Lagrangian is: $\mathcal{L}_{\rho} = \sum_{i=1}^b \Big( \sum_{j=1}^n \big( \theta_j(\b{Y}_i) \big) + h(\b{V}_i) \Big) + \textbf{tr}\big(\b{\Lambda}^\top (\b{Y} - \b{V})\big) + (\rho/2)\, ||\b{Y} - \b{V}||_F^2 =  \sum_{i=1}^b \Big( \sum_{j=1}^n \big( \theta_j(\b{Y}_i) \big) + h(\b{V}_i) \Big) + (\rho/2)\, ||\b{Y} - \b{V} + \b{J}||_F^2 - (\rho/2)\, ||\b{\Lambda}||_F^2$, where $\b{\Lambda} := \cup_{i=1}^b \b{\Lambda}_i$ is the Lagrange multiplier, $\rho > 0$, and $\b{J} := (1/\rho) \b{\Lambda} = (1/\rho) \cup_{i=1}^b \b{\Lambda}_i = \cup_{i=1}^b \b{J}_i$. The term $(\rho/2)\, ||\b{\Lambda}||_F^2$ is a constant with respect to $\b{Y}$ and $\b{V}$ and can be dropped. 
The updates of $\b{Y}$, $\b{V}$, and $\b{J}$ are done as \cite{boyd2011distributed,otero2018alternate}:
\begin{align}
\b{Y}_i^{(\nu+1)} & := \arg \min_{\b{Y}_i} \Big( \sum_{j=1}^n \big( \theta_j(\b{Y}_i) \big) + (\rho/2)\, ||\b{Y}_i - \b{V}_i^{(\nu)} + \b{J}_i^{(\nu)}||_F^2 \Big), \label{equation_LLISE_ADMM_Y_update} \\
\b{V}_i^{(\nu+1)} & := \arg \min_{\b{V}_i} \Big( h_2(\b{V}_i) + (\rho/2)\, ||\b{Y}_i^{(\nu+1)} - \b{V}_i + \b{J}_i^{(\nu)}||_F^2 \Big), \label{equation_LLISE_ADMM_V_update} \\
\b{J}^{(\nu+1)} & := \b{J}^{(\nu)} + \b{Y}^{(\nu+1)} - \b{V}^{(\nu+1)}. \label{equation_LLISE_ADMM_J_update}
\end{align}
The gradient of the objective function in Eq. (\ref{equation_LLISE_ADMM_Y_update}) is $\sum_{j=1}^n \big(\nabla \theta_j(\b{Y}_i)\big) + \rho\, (\b{Y}_i - \b{V}_i^{(\nu)} + \b{J}_i^{(\nu)})$. Similar to Eq. (\ref{equation_LLISE_reconst_ADMM_w_update}), we replace Eq. (\ref{equation_LLISE_ADMM_Y_update}) with one iteration of gradient descent. 

With the same explanation for Eq. (\ref{equation_LLISE_reconst_ADMM_xi_update}), Eq. (\ref{equation_LLISE_ADMM_V_update}) is equivalent to the projection $\mathrm{\Pi}(\b{Y}_i^{(\nu+1)} + \b{J}_i^{(\nu)})$.
One of the constraints in Eq. (\ref{equation_LLISE_linearEmbedding_2}) is $\b{Y}_i^\top \b{1} = \b{0}$. Therefore, the row mean of the matrix should removed, i.e., $\b{Y}_i := \b{H} \b{Y}_i$, where $\mathbb{R}^{n \times n} \ni \b{H} := \b{I} - (1/n) \b{1}\b{1}^\top$ is the centering matrix and $\b{1}$ is the vector of ones.
The other constraint in Eq. (\ref{equation_LLISE_linearEmbedding_2}) is $(1/n)\b{Y}_i^\top \b{Y}_i = \b{I}$.
The variable of proximal operator, which is a projection here, is a matrix and not a vector. 
According to \cite{parikh2014proximal}, if $F$ is a convex and orthogonally invariant function and it works on the singular values of a matrix variable $\b{A} \in \mathbb{R}^{n \times p}$, i.e., $F = f \circ \sigma$ where the function $\sigma(\b{A})$ gives the vector of singular values of $\b{A}$, then the proximal operator is $\textbf{prox}_{\lambda, F}(\b{A}) := \b{Q}\,\, \textbf{diag}\Big(\textbf{prox}_{\lambda, f}\big(\sigma(\b{A})\big)\Big)\,\, \b{\Omega}^\top$.
The $\b{Q} \in \mathbb{R}^{n \times p}$ and $\b{\Omega} \in \mathbb{R}^{p \times p}$ are the matrices of left and right singular vectors of $\b{A}$, respectively.
In our constraint $(1/n)\b{Y}_i^\top \b{Y}_i = \b{I}$, the function $F$ deals with the singular values of $\b{Y}_i$. The reason is that we want: $\b{Y}_i \overset{\text{SVD}}{=} \b{Q} \b{\Sigma} \b{\Omega}^\top \implies (1/n)\b{Y}_i^\top \b{Y}_i = (1/n) \b{\Omega} \b{\Sigma} \b{Q}^\top \b{Q} \b{\Sigma} \b{\Omega}^\top \overset{(a)}{=} (1/n) \b{\Omega} \b{\Sigma}^2 \b{\Omega}^\top \overset{\text{set}}{=} \b{I} \implies (1/n) \b{\Omega} \b{\Sigma}^2 \b{\Omega}^\top \b{\Omega} = \b{\Omega} \overset{(b)}{\implies} (1/n) \b{\Omega} \b{\Sigma}^2 = \b{\Omega} \implies \b{\Sigma} = n \b{I}$, where $(a)$ and $(b)$ are because $\b{Q}$ and $\b{\Omega}$ are orthogonal matrices. Thus, projection onto the second constraint is equivalent to decomposing the matrix with Singular Value Decomposition (SVD) and setting all the singular values to $n$.
To sum up, $\mathrm{\Pi}(\b{Y}_i^{(\nu+1)} + \b{J}_i^{(\nu)})$ first removes the row mean of $(\b{Y}_i^{(\nu+1)} + \b{J}_i^{(\nu)})$ and then sets the singular values of $(\b{Y}_i^{(\nu+1)} + \b{J}_i^{(\nu)})$  to $n$.
In summary, the Eqs. (\ref{equation_LLISE_ADMM_Y_update}), (\ref{equation_LLISE_ADMM_V_update}), and (\ref{equation_LLISE_ADMM_J_update}) can be restated as:
\begin{equation}\label{equation_LLISE_ADMM_updates}
\begin{aligned}
\b{Y}_i^{(\nu+1)} & := \b{Y}_i^{(\nu)} - \eta\, \sum_{j=1}^n \big(\nabla \theta_j(\b{Y}_i)\big) - \eta\,\rho\, (\b{Y}_i - \b{V}_i^{(\nu)} + \b{J}_i^{(\nu)}), \\
\b{V}_i^{(\nu+1)} & := \mathrm{\Pi}(\b{Y}_i^{(\nu+1)} + \b{J}_i^{(\nu)}), \\
\b{J}^{(\nu+1)} & := \b{J}^{(\nu)} + \b{Y}^{(\nu+1)} - \b{V}^{(\nu+1)}.
\end{aligned}
\end{equation}
Iteratively solving Eq. (\ref{equation_LLISE_ADMM_updates}) until convergence gives us the $\b{Y}_i$ for the image blocks indexed by $i$. The rows of $\b{Y}_i$ are the $p$-dimensional embedded image blocks in the \textit{LLISE manifold}.  
Unlike LLE, the first column of $\b{Y}_i$ is not ignored in LLISE because it is not based on $\ell_2$ norm and thus eigenvalue problem.

\subsection{Embedding The Out-of-sample Data}

There exist two methods in the literature for extension of LLE for out-of-sample embedding. The first method is based on the concept of eigenfunctions \cite{bengio2004out} and the second method uses linear reconstruction of the out-of-sample data \cite{saul2003think}. The first method cannot be used for LLISE because it does not result in closed-form eigenvalue problem as in LLE. We use the second approach.

Suppose we have $n_t$ out-of-sample images and $\breve{\b{x}}_{j,i}^{(t)}$ denotes the $i$-th block in the $j$-th out-of-sample image.
For the $i$-th block in every out-of-sample image, we first find the $k$-NN among the $i$-th block in training images. 
Let $_r\breve{\b{x}}_{j,i}^{(t)}$ and $\mathbb{R}^{q \times k} \ni \breve{\b{X}}_{j,i}^{(t)} := [\,_1\breve{\b{x}}_{j,i}^{(t)}, \dots, \,_k\breve{\b{x}}_{j,i}^{(t)}]$ denote the $r$-th training neighbor of $\breve{\b{x}}_{j,i}^{(t)}$ and 
the matrix including the training neighbors of $\breve{\b{x}}_{j,i}^{(t)}$, respectively.
We want to reconstruct every out-of-sample image block by its training neighbors:
\begin{equation}\label{equation_LLISE_outOfSample}
\begin{aligned}
& \underset{\widetilde{\b{W}}_i^{(t)}}{\text{minimize}}
& & \sum_{i=1}^b \varepsilon(\widetilde{\b{W}}_i^{(t)}) := \sum_{i=1}^b \sum_{j=1}^{n_t} \Big|\Big|\, \breve{\b{x}}_{j,i}^{(t)} - \sum_{r=1}^k \,_r\widetilde{w}_{j,i}^{(t)}\, \,_r\breve{\b{x}}_{j,i}^{(t)}\Big|\Big|_S, \\
& \text{subject to}
& & \sum_{r=1}^k (_r\widetilde{w}_{j,i}^{(t)})^2 = 1, ~~~ \forall i \in \{1, \dots, b\}, ~~ \forall j \in \{1, \dots, n_t\},
\end{aligned}
\end{equation}
where $\mathbb{R}^{n_t \times k} \ni \widetilde{\b{W}}_i^{(t)} := [\widetilde{\b{w}}_{1,i}^{(t)}, \dots, \widetilde{\b{w}}_{n_t,i}^{(t)}]^\top$ includes the weights, $\mathbb{R}^k \ni \widetilde{\b{w}}_{j,i}^{(t)} := [\,_1\widetilde{w}_{j,i}^{(t)}, \dots, \,_k\widetilde{w}_{j,i}^{(t)}]^\top$ includes the weights of linear reconstruction of the $i$-th block in the $j$-th out-of-sample image using the $i$-th block in its $k$ training neighbors.
Note that Eq. (\ref{equation_LLISE_outOfSample}) is similar to Eq. (\ref{equation_LLISE_linearReconstruct}) and is solved using Eq. (\ref{equation_LLISE_reconst_ADMM_updates}) where $\widetilde{\b{w}}_{j,i}^{(t)}$, $\breve{\b{x}}_{j,i}^{(t)}$, and $\breve{\b{X}}_{j,i}^{(t)}$ are used in the expressions.

The embedding $\b{y}_{j,i}^{(t)}$ of the $i$-th block in the $j$-th out-of-sample image, i.e., $\b{x}_{j,i}^{(t)}$, is obtained by the linear reconstruction of the embedding of the $i$-th block in its $k$ training neighbors:
\begin{align}\label{equation_LLISE_outOfSample_embedding}
\mathbb{R}^p \ni \b{y}_{j,i}^{(t)} = \sum_{r=1}^{k} \,_r\widetilde{w}_{j,i}^{(t)}\, _r\b{y}_{j,i}^{(t)},
\end{align}
where $_r\b{y}_{j,i}^{(t)} \in \mathbb{R}^p$ is the embedding of $_r\breve{\b{x}}_{j,i}^{(t)}$ which was found by the linear embedding of the training data, $\b{Y}_i$.

\section{Kernel Locally Linear Image Structural Embedding}

We can map the block $\breve{\b{x}}_i \in \mathbb{R}^d$ to higher-dimensional feature space hoping to have the data fall close to a simpler-to-analyze manifold in the feature space. Suppose $\b{\phi}: \breve{\b{x}} \rightarrow \mathcal{H}$ is a function which maps the data $\breve{\b{x}}$ to the feature space. In other words, $\breve{\b{x}} \mapsto \b{\phi}(\breve{\b{x}})$. Let $t$ denote the dimensionality of the feature space, i.e., $\b{\phi}(\breve{\b{x}}) \in \mathbb{R}^t$. We usually have $t \gg d$. 
The kernel of the $i$-th block in images $1$ and $2$, which are $\breve{\b{x}}_{1,i}$ and $\breve{\b{x}}_{2,i}$, is $k(\breve{\b{x}}_{1,i}, \breve{\b{x}}_{2,i}) := \b{\phi}(\breve{\b{x}}_{1,i})^\top \b{\phi}(\breve{\b{x}}_{2,i}) \in \mathbb{R}$. 

Let $\b{K} = \b{\Phi}(\breve{\b{X}}_i)^\top \b{\Phi}(\breve{\b{X}}_i) \in \mathbb{R}^{n \times n}$ be the kernel between the $i$-th block in the $n$ images. 
We can normalize it as $\b{K}(a,b) := \b{K}(a,b) / \sqrt{\b{K}(a,a) \b{K}(b,b)}$ where $\b{K}(a,b)$ denotes the $(a,b)$-th element of the kernel matrix \cite{ah2010normalized}. Then, the kernel is double-centered as $\b{K} := \b{H} \b{K} \b{H}$.
The reason for double-centering is that Eq. (\ref{equation_SSIM_distance}) requires $\b{\phi}(\breve{\b{x}}_i)$ and thus the $\b{\Phi}(\breve{\b{X}}_i)$ to be centered. 
Therefore, in kernel LLISE, we center the kernel rather than centering $\breve{\b{x}}_i$.
Kernel LLISE maps the data to the feature space and performs the steps of $k$-NN and linear reconstruction in the feature space.

% The kernel matrix for the $i$-th block among the $n$ images is $\mathbb{R}^{n \times n} \ni \b{K}_i := \b{\Phi}(\breve{\b{X}}_i)^\top \b{\Phi}(\breve{\b{X}}_i)$ where $\b{\Phi}(\breve{\b{X}}_i) := [\b{\phi}(\breve{\b{x}}_{1,i}), \dots, \b{\phi}(\breve{\b{x}}_{n,i})] \in \mathbb{R}^{t \times n}$.

\subsection{Embedding The Training Data}

\subsubsection{$k$-Nearest Neighbors}

The Euclidean distance in the feature space is \cite{scholkopf2001kernel}:
\begin{align}
||\b{\phi}(\breve{\b{x}}_{a,i}) - \b{\phi}(\breve{\b{x}}_{b,i})||_2 = \sqrt{k(\breve{\b{x}}_{a,i}, \breve{\b{x}}_{a,i}) -2 k(\breve{\b{x}}_{a,i}, \breve{\b{x}}_{b,i}) + k(\breve{\b{x}}_{b,i}, \breve{\b{x}}_{b,i})}.
\end{align}
For every block $i$ amongst the images, we construct the $k$-NN graph using the distances of the blocks in the feature space.
Therefore, every block has $k$ neighbors in the feature space. Let the matrix $\mathbb{R}^{t \times k} \ni \b{\Phi}(\breve{\b{X}}_{j,i}) := [\b{\phi}(_1\breve{\b{x}}_{j,i}), \dots, \b{\phi}(_k\breve{\b{x}}_{j,i})]$ include the neighbors of $\breve{\b{x}}_{j,i}$ in the feature space.

\subsubsection{Linear Reconstruction by the Neighbors}

For finding the reconstruction weights $\mathbb{R}^k \ni \widetilde{\b{w}}_{j,i} = [_1\widetilde{w}_{j,i}, \dots, \,_k\widetilde{w}_{j,i}]^\top$, the Eq. (\ref{equation_LLISE_linearReconstruct}) is used in the feature space:
\begin{equation}\label{equation_kernel_LLISE_linearReconstruct}
\begin{aligned}
& \underset{\widetilde{\b{W}}_i}{\text{minimize}}
& & \varepsilon(\widetilde{\b{W}}_i) := \sum_{i=1}^b \sum_{j=1}^n \Big|\Big|\b{\phi}(\breve{\b{x}}_{j,i}) - \sum_{r=1}^k \,_r\widetilde{w}_{j,i} \, \b{\phi}(_r\breve{\b{x}}_{j,i})\Big|\Big|_S, \\
& \text{subject to}
& & \sum_{r=1}^k \,_r\widetilde{w}_{j,i}^2 = 1, ~~~ \forall i \in \{1, \dots, b\}, ~~ \forall j \in \{1, \dots, n\}.
\end{aligned}
\end{equation}
Let $f^{\phi}(\b{\widetilde{w}}_{j,i}) := \big|\big|\b{\phi}(\breve{\b{x}}_{j,i}) - \sum_{r=1}^k \,_r\widetilde{w}_{ij} \, \b{\phi}(_r\breve{\b{x}}_{j,i})\big|\big|_S$.
According to Eq. (\ref{equation_SSIM_distance}), we have:
\begin{align}\label{equation_kernel_LLISE_f}
\mathbb{R} \ni f^{\phi}(\widetilde{\b{w}}_{j,i}) = \frac{k_{j,i} + \widetilde{\b{w}}_{j,i}^\top\, \b{K}_{j,i}\, \widetilde{\b{w}}_{j,i} - 2\, \widetilde{\b{w}}_{j,i}^\top\, \b{k}_{j,i}}{k_{j,i} + \widetilde{\b{w}}_{j,i}^\top\, \b{K}_{j,i}\, \widetilde{\b{w}}_{j,i} + c},
\end{align}
where $\mathbb{R} \ni k_{j,i} :=  \b{\phi}(\breve{\b{x}}_{j,i})^\top \b{\phi}(\breve{\b{x}}_{j,i})$, $\mathbb{R}^k \ni \b{k}_{j,i} :=  \b{\Phi}(\breve{\b{X}}_{j,i})^\top \b{\phi}(\breve{\b{x}}_{j,i})$, and $\mathbb{R}^{k \times k} \ni \b{K}_{j,i} :=  \b{\Phi}(\breve{\b{X}}_{j,i})^\top \b{\Phi}(\breve{\b{X}}_{j,i})$.
The gradient of $f^{\phi}(\widetilde{\b{w}}_{j,i})$ with respect to $\widetilde{\b{w}}_{j,i}$ is:
\begin{align}\label{equation_kernel_LLISE_derivative_f}
\mathbb{R}^k \ni \nabla f^{\phi}(\widetilde{\b{w}}_{j,i}) = \frac{2\, \Big(\big(1 - f^{\phi}(\widetilde{\b{w}}_{j,i})\big) \b{K}_{j,i}\, \widetilde{\b{w}}_{j,i} - \b{k}_{j,i}\Big)}{k_{j,i} + \widetilde{\b{w}}_{j,i}^\top\, \b{K}_{j,i}\, \widetilde{\b{w}}_{j,i} + c}.
\end{align}
We can use Eq. (\ref{equation_LLISE_reconst_ADMM_updates}) for solving Eq. (\ref{equation_kernel_LLISE_linearReconstruct}) where $\nabla f^{\phi}(\widetilde{\b{w}}_{j,i})$ is used in place of $\nabla f(\widetilde{\b{w}}_{j,i})$.
The linear embedding in kernel LLISE is the same as the linear embedding in LLISE. The rows of obtained $\b{Y}_i$ are the $i$-th embedded block of the images in \textit{kernel LLISE manifold}.

\subsection{Embedding The Out-of-sample Data}

For embedding every out-of-sample image, we reconstruct it by its training neighbors in the feature space.
The Eq. (\ref{equation_LLISE_outOfSample}) in the feature space is:
\begin{equation}\label{equation_kernel_LLISE_outOfSample}
\begin{aligned}
& \underset{\widetilde{\b{W}}_i^{(t)}}{\text{minimize}}
& & \sum_{i=1}^b \varepsilon(\widetilde{\b{W}}_i^{(t)}) := \sum_{i=1}^b \sum_{j=1}^{n_t} \Big|\Big|\, \b{\phi}(\breve{\b{x}}_{j,i}^{(t)}) - \sum_{r=1}^k \,_r\widetilde{w}_{j,i}^{(t)}\, \b{\phi}(_r\breve{\b{x}}_{j,i}^{(t)})\Big|\Big|_S, \\
& \text{subject to}
& & \sum_{r=1}^k (_r\widetilde{w}_{j,i}^{(t)})^2 = 1, ~~~ \forall i \in \{1, \dots, b\}, ~~ \forall j \in \{1, \dots, n_t\},
\end{aligned}
\end{equation}
which is similar to Eq. (\ref{equation_kernel_LLISE_linearReconstruct}) and is solved similarly.  
Here, the used kernels are $\mathbb{R} \ni k_{j,i} =  \b{\phi}(\breve{\b{x}}_{j,i}^{(t)})^\top \b{\phi}(\breve{\b{x}}_{j,i}^{(t)})$, $\mathbb{R}^k \ni \b{k}_{j,i} =  \b{\Phi}(\breve{\b{X}}_{j,i}^{(t)})^\top \b{\phi}(\breve{\b{x}}_{j,i}^{(t)})$, and $\mathbb{R}^{k \times k} \ni \b{K}_{j,i} =  \b{\Phi}(\breve{\b{X}}_{j,i}^{(t)})^\top \b{\Phi}(\breve{\b{X}}_{j,i}^{(t)})$.
After finding the weights $\widetilde{\b{w}}_{j,i}^{(t)} = [\,_1\widetilde{w}_{j,i}^{(t)}, \dots, \,_k\widetilde{w}_{j,i}^{(t)}]$ from Eq. (\ref{equation_kernel_LLISE_outOfSample}), the embedding of the out-of-sample $\b{x}_{j,i}^{(t)}$ is found using Eq. (\ref{equation_LLISE_outOfSample_embedding}) where $_r\b{y}_{j,i}^{(t)} \in \mathbb{R}^p$ is the embedding of $_r\breve{\b{x}}_{j,i}^{(t)}$ in kernel LLISE.

\section{Experiments}

\begin{figure}[!t]
\centering
\includegraphics[width=4.7in]{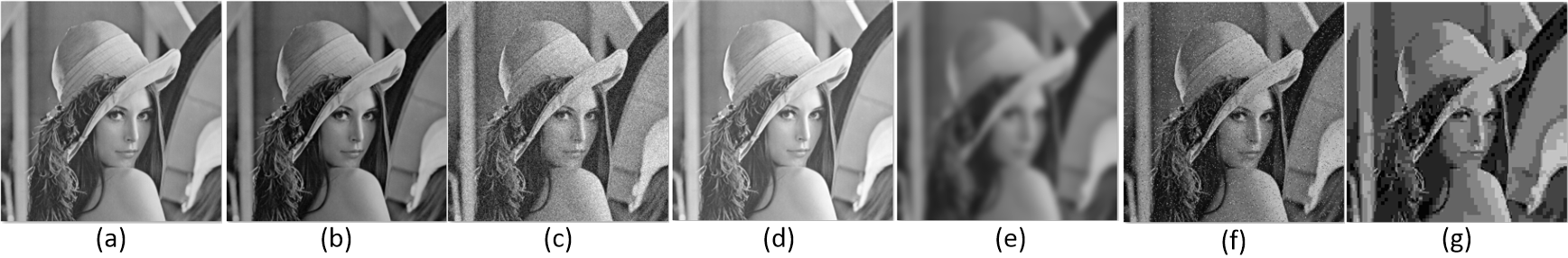}
\caption{Samples from the training dataset: (a) original image, (b) contrast stretched, (c) Gaussian noise, (d) luminance enhanced, (e) Gaussian blurring, (f) salt \& pepper impulse noise, and (g) JPEG distortion.}
\label{fig_training_dataset}
\end{figure}

\textbf{Training Dataset:}
We made a dataset out of the standard \textit{Lena} image. Six different types of distortions were applied on the original \textit{Lena} image (see Fig. \ref{fig_training_dataset}), each of which has $20$ images in the dataset with different MSE values. Therefore, the size of the training set is $121$ including the original image. 
For every type of distortion, $20$ different levels of MSE, i.e., from $\text{MSE} = 45$ to $\text{MSE} = 900$ with step $45$, were generated to have images on the equal-MSE or \textit{iso-error} hypersphere \cite{wang2006modern}. 

\begin{figure}[!t]
\centering
\includegraphics[width=4.8in]{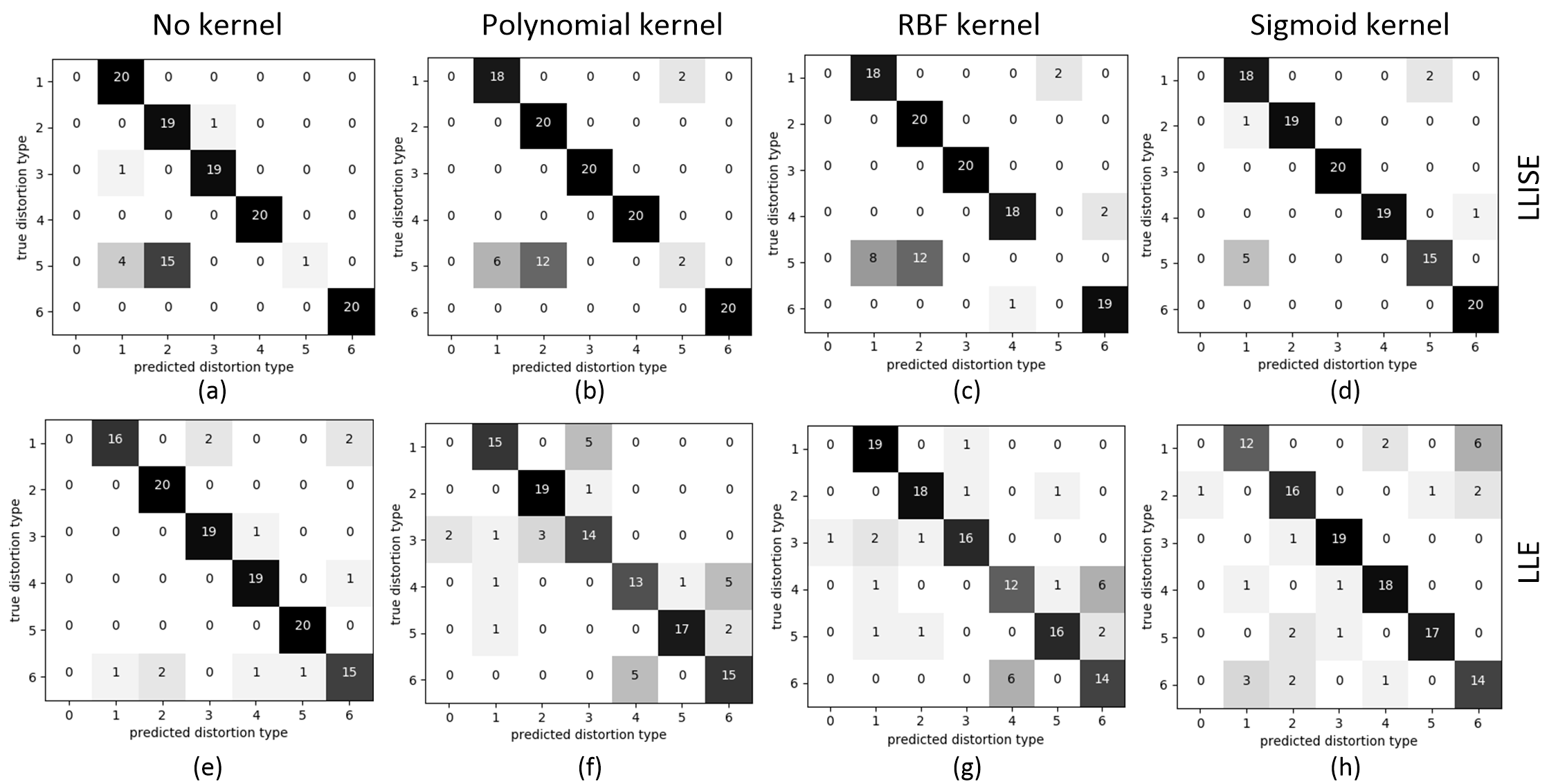}
\caption{Confusion matrices for recognition of distortion types with a 1NN classifier used in the embedded space. Matrices (a) and (e) correspond to LLISE and LLE, respectively. Matrices (b) to (d) are for kernel LLISE and (f) and (g) are for kernel LLE with polynomial, RBF, and sigmoid kernels, respectively. The $0$ label corresponds to the original image and the labels $1$ to $6$ are the distortion types with the same order as in Fig. \ref{fig_training_dataset}.}
\label{fig_confusion_matrices}
\end{figure}

\BlankLine
\noindent
\textbf{Embedding the Training Images:}
We embedded the blocks in the training images.
In $k$-NN, we used $k=10$.
For linear reconstruction, we used $\rho=\eta=0.1$ in LLISE and $10\rho=\eta=0.1$ in kernel LLISE. For linear embedding, we used $\rho=\eta=0.01$.
We took $q= 64$ ($8 \times 8$ blocks inspired by \cite{otero2014unconstrained,otero2018alternate}), $p=4$, and $d=512\times 512 = 262144$.
In order to evaluate the obtained embedded manifold, we used the 1-Nearest Neighbor (1NN) classifier to recognize the distortion type of every block. The 1NN is useful to show how to evaluate the manifold by closeness of the embedded distortions.  
The distortion type of an image comes from a majority vote among the blocks. 
The polynomial $(\gamma\, \breve{\b{x}}_1^\top \breve{\b{x}}_2 + 1)^3$, Radial Basis function (RBF) $\exp(-\gamma\, ||\breve{\b{x}}_1 - \breve{\b{x}}_2||_2^2)$, and sigmoid $\tanh (\gamma\, \breve{\b{x}}_1^\top \breve{\b{x}}_2 + 1)$ kernels were tested for kernel LLISE, where $\gamma := 1/q$.
The confusion matrices for distortion recognition are shown in Fig. \ref{fig_confusion_matrices}. 
Also, the LLISE and kernel LLISE are compared with LLE and kernel LLE in this figure. Except for impulse noise, LLISE and kernel LLISE had better performance compared to LLE and kernel LLE. In other distortions, especially in JPEG distortion and contrast stretch, the performances of LLE and kernel LLE are not acceptable because LLE uses $\ell_2$ norm rather than SSIM distance. 

\begin{figure}[!t]
\centering
\includegraphics[width=3.5in]{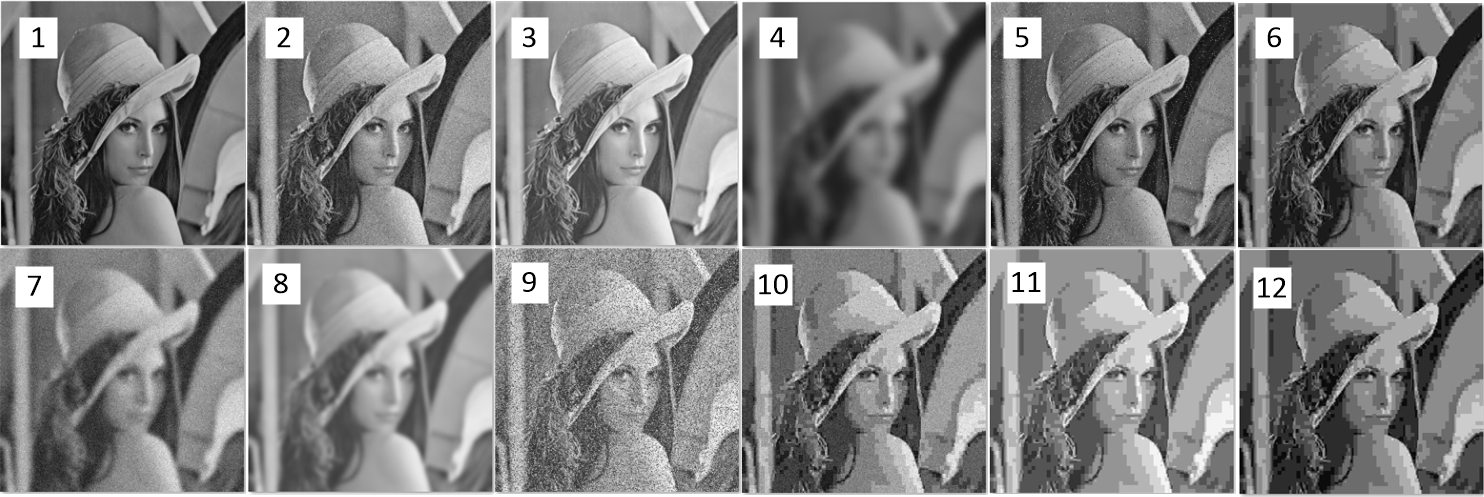}
\caption{Out-of-sample images with different types of distortions having $\text{MSE}=500$: (1) stretching contrast, (2) Gaussian noise, (3) luminance enhancement, (4) Gaussian blurring, (5) impulse noise, (6) JPEG distortion, (7) Gaussian blurring $+$ Gaussian noise, (8) Gaussian blurring $+$ luminance enhancement, (9) impulse noise $+$ luminance enhancement, (10) JPEG distortion $+$ Gaussian noise, (11) JPEG distortion $+$ luminance enhancement, and (12) JPEG distortion $+$ stretching contrast.}
\label{fig_test_dataset}
\end{figure}

\BlankLine
\noindent
\textbf{Out-of-sample Embedding:}
For out-of-sample embedding, we made $12$ test images with $\text{MSE}=500$ having different distortions and some having a combination of different distortions (see Fig. \ref{fig_test_dataset}).
Again, for linear reconstruction, we used $\rho=\eta=0.1$ in LLISE and $10\rho=\eta=0.1$ in kernel LLISE. 
The same 1NN classification was done for the test images. Table \ref{table_outOfSample_recongition} reports the top two votes of blocks for every image with the percentage of blocks voting for those distortions. This table also shows the recognition of distortions using LLE and kernel LLE. Note that LLE does not perform block-wise and thus it has only one recongnition label for the whole image. 
As expected for LLISE and kernel LLISE, in most cases, at least one of the two top votes recognized the type of distortion(s) the out-of-sample images had. However, LLE and kernel LLE performed poorly on the out-of-sample images.

\begin{table*}[!t]
\begin{minipage}{\textwidth}
\caption{Recognition of distortions for out-of-sample images. Letters O, C, G, L, B, I, and J correspond to original image, contrast stretch, Gaussian noise, luminance enhanced, blurring, impulse noise, and JPEG distortion, respectively.}
\label{table_outOfSample_recongition}
\setlength\extrarowheight{5pt}
\centering
\scalebox{0.6}{    %%% --> for resizing tables
\begin{tabular}{l || c | c | c | c | c | c | c | c | c | c | c | c}
\hline
image &  1 & 2 & 3 & 4 & 5 &  6 & 7 & 8 & 9 & 10 & 11 & 12 \\
\hline
\hline
distortion & C & G & L & B & I & J & B $+$ G & B $+$ L & I $+$ L & J $+$ G & J $+$ L & J $+$ C \\
\hline
\hline
\multirow{2}{*}{LLISE}  
& 42.9\% C & 42.2\% G & 35.6\% L & 44.5\% B & 31.2\% G & 43.2\% J & 46.8\% G & 41.3\% B & 55.9\% G & 33.5\% G & 39.6\% J & 40.7\% J \\
& 22.8\% L & 29.3\% I & 29.6\% C & 15.7\% J & 28.4\% I & 16.8\% B & 34.4\% I & 14.7\% J & 39.6\% I & 26.5\% I & 16.3\% B & 16.3\% L \\
\hline
\multirow{2}{*}{kernel LLISE (polynomial)} 
& 69.7\% C & 23.7\% L & 97.5\% L & 74.3\% B & 45.4\% C & 76.6\% J & 20.7\% G & 78.1\% L & 35.1\% G & 27.7\% L & 76.9\% L & 45.1\% J \\
& 14.7\% I & 21.3\% G & 0.8\% C & 14.5\% J & 19.5\% I & 13.9\% B & 17.9\% B & 5.7\% I & 23.5\% L & 16.6\% G & 7.0\% B & 28.6\% B \\
\hline
\multirow{2}{*}{kernel LLISE (RBF)} 
& 62.1\% C & 25.6\% L & 72.6\% L & 58.1\% B & 42.3\% C & 59.6\% J & 23.7\% B & 58.1\% L & 33.3\% L & 26.5\% L & 57.8\% L & 40.0\% J \\
& 12.0\% I & 16.1\% B & 9.5\% C & 16.3\% J & 17.0\% I & 16.4\% B & 18.5\% J & 11.9\% B & 24.7\% G & 19.6\% J & 13.0\% B & 24.8\% B \\
\hline
\multirow{2}{*}{kernel LLISE (sigmoid)} 
& 63.0\% C & 28.5\% L & 92.4\% L & 68.5\% B & 51.9\% C & 55.6\% J & 39.2\% B & 77.8\% L & 57.0\% L & 31.1\% L & 76.1\% L & 42.8\% J \\
& 14.5\% I & 24.5\% C & 3.5\% B & 15.5\% J & 14.0\% I & 19.7\% B & 19.5\% L & 10.7\% B & 12.6\% C & 24.6\% B & 10.8\% B & 29.4\% B \\
\hline
\hline
LLE & C & L & L & B & C & J & B & C & C & L & L & B \\
\hline
kernel LLE (polynomial) & L & C & L & B & C & J & B & L & L & B & L & J \\
\hline
kernel LLE (RBF) & L & C & C & J & C & J & B & L & L & B & L & J \\
\hline
kernel LLE (sigmoid) & C & L & L & B & C & J & J & C & L & C & C & C \\
\hline
\end{tabular}%
}
\end{minipage}
\end{table*}

\section{Conclusion and Future Work}

This paper introduced the concept of an image structure manifold which discriminates the types of distortions applied on the images and captures the structure of an image. A new method, named LLISE, was proposed for learning this manifold in both original and feature spaces. The LLISE is inspired by LLE which uses $\ell_2$ norm.
As a possible future work, we seek to design other new methods for learning the image structure manifold.

\bibliographystyle{splncs}      % basic style, author-year citations
\bibliography{references.bib}            % name your BibTeX data base

\section{Supplementary Material: Review of Locally Linear Embedding}

In the paper, we did not completely review the details of LLE \cite{roweis2000nonlinear} for the sake of brevity. Here, we review it with a little more details. We do not mention the derivations of equations in LLE because it is out of the scope of this paper.

\subsection{Embedding The Training Data}

\subsubsection{$k$-Nearest Neighbors}

In LLE \cite{roweis2000nonlinear}, first a $k$-Nearest Neighbor ($k$-NN) graph is found using pairwise Euclidean distances. Every data point $\b{x}_j \in \mathbb{R}^d$ is reconstructed by its $k$ neighbors $\mathbb{R}^{d \times k} \ni \b{X}_{j} := [ \,_1\b{x}_{j}, \dots,\, _k\b{x}_{j}]$ where $_r\b{x}_{j}$ denotes the $r$-th neighbor of $\b{x}_j$. 
\subsubsection{Linear Reconstruction by the Neighbors}

If $\mathbb{R}^k \ni \widetilde{\b{w}}_{j} := [\,_1\widetilde{w}_{j}, \dots, \,_k\widetilde{w}_{j}]^\top$ denotes the reconstruction weights for the $\b{x}_j$ using its $k$ neighbors, the reconstruction problem with the weights adding to one is: 
\begin{equation}\label{equation_LLE_linearReconstruct}
\begin{aligned}
& \underset{\widetilde{\b{w}}_j}{\text{minimize}}
& & \sum_{j=1}^n \Big|\Big|\b{x}_j - \sum_{r=1}^k \,_r\widetilde{w}_{j} \,_r\b{x}_{j}\Big|\Big|_2^2, \\
& \text{subject to}
& & \sum_{r=1}^k \,_r\widetilde{w}_{j} = 1, ~~~ \forall j \in \{1, \dots, n\}.
\end{aligned}
\end{equation}
The solution to the Eq. (\ref{equation_LLE_linearReconstruct}) is:
\begin{align}
\widetilde{\b{w}}_j = \frac{\lambda_j}{2} \b{G}_j^{-1} \b{1} = \frac{\b{G}_j^{-1} \b{1}}{\b{1}^\top \b{G}_j^{-1} \b{1}},
\end{align}
where $\mathbb{R}^{k \times k} \ni \b{G}_j := (\b{x}_j \b{1}^\top - \b{X}_j)^\top (\b{x}_j \b{1}^\top - \b{X}_j)$.

\subsubsection{Linear Embedding}

If $\b{y}_j \in \mathbb{R}^p$ denotes the embedded $j$-th data point, the embedding problem with unit covariance is: 
\begin{equation}\label{equation_LLE_linearEmbedding}
\begin{aligned}
& \underset{\widetilde{\b{w}}_j}{\text{minimize}}
& & \sum_{j=1}^n \Big|\Big|\b{y}_j - \sum_{r=1}^n \,_rw_{j}\, \b{y}_r\Big|\Big|_2^2, \\
& \text{subject to}
& & \frac{1}{n} \sum_{j=1}^n \b{y}_j \b{y}_j^\top = \b{I}, ~~~ \sum_{j=1}^n \b{y}_j = \b{0}, ~~~ \forall j \in \{1, \dots, n\},
\end{aligned}
\end{equation}
where $_rw_{j}$ is the weight obtained from the linear reconstruction if $\b{x}_r$ is a neighbor of $\b{x}_j$ and zero otherwise. Let $\mathbb{R}^n \ni \b{w}_i := [w_{i1}, \dots, w_{in}]^\top$ and $\mathbb{R}^{n \times n} \ni \b{W} := [\b{w}_1, \dots, \b{w}_n]^\top$.

The solution to Eq. (\ref{equation_LLE_linearEmbedding}) is:
\begin{align}
\b{M}\b{Y} = \b{Y} (\frac{1}{n}\b{\Lambda}), \label{equation_LLE_linearEmbedding_eigenproblem}
\end{align}
which is the eigenvalue problem for $\b{M}$. Note that $\mathbb{R}^{n \times p} \ni \b{Y} := [\b{y}_1, \dots, \b{y}_n]^\top$ and $\mathbb{R}^{n \times n} \ni \b{M} := (\b{I} - \b{W})^\top (\b{I} - \b{W})$.
Therefore, the columns of $\b{Y}$ are the eigenvectors of $\b{M}$ where eigenvalues are the diagonal elements of $(1/n)\b{\Lambda}$.

The $\b{M} = (\b{I} - \b{W})^\top (\b{I} - \b{W})$. The $(\b{I} - \b{W})$ is the Laplacian matrix for $\b{W}$ because the columns of $\b{W}$, which are $\b{w}_j$'s, add to one (because of the constraint used in Eq. (\ref{equation_LLE_linearReconstruct})).
As the $k$-nearest neighbor graph, or $\b{W}$, is a connected graph, $(\b{I} - \b{W})$ has one zero eigenvalue whose eigenvector is $\b{1} = [1, 1, \dots, 1]^\top$. After sorting the eigenvectors from smallest to largest eigenvalues, we ignore the first eigenvector having zero eigenvalue and take the $p$ eigenvectors of $\b{M}$ with non-zero eigenvalues as the columns of $\b{Y} \in \mathbb{R}^{n \times p}$.

\subsection{Embedding The Out-of-sample Data}

One way of extending LLE for out-of-sample data point is using linear reconstruction \cite{saul2003think}.
For every out-of-sample data point $\b{x}_j^{(t)}$, we first find the $k$ nearest neighbors among the training points. 
Let $_r\b{x}_{j}^{(t)}$ denote the $r$-th training neighbor of $\b{x}_j^{(t)}$ and let the matrix $\mathbb{R}^{d \times k} \ni \b{X}_j^{(t)} := [\,_1\b{x}_{j}^{(t)}, \dots, \,_k\b{x}_{j}^{(t)}]$ include the training neighbors of $\b{x}_j^{(t)}$.
We want to reconstruct every out-of-sample point by its training neighbors:
\begin{equation}\label{equation_outOfSample_LLE_linearReconstruction}
\begin{aligned}
& \underset{\widetilde{\b{w}}_j^{(t)}}{\text{minimize}}
& & \sum_{j=1}^{n_t} \Big|\Big|\b{x}_j^{(t)} - \sum_{r=1}^k \,_r\widetilde{w}_{j}^{(t)} \,_r\b{x}_{j}^{(t)}\Big|\Big|_2^2, \\
& \text{subject to}
& & \sum_{r=1}^k \,_r\widetilde{w}_{j}^{(t)} = 1, ~~~ \forall j \in \{1, \dots, n_t\},
\end{aligned}
\end{equation}
where $\mathbb{R}^k \ni \widetilde{\b{w}}_j^{(t)} := [\,_1\widetilde{w}_{j}^{(t)}, \dots, \,_k\widetilde{w}_{j}^{(t)}]^\top$ includes the weights of linear reconstruction of the $j$-th out-of-sample data point using its $k$ training neighbors.

The Eq. (\ref{equation_outOfSample_LLE_linearReconstruction}) is similar to Eq. (\ref{equation_LLE_linearReconstruct}) and thus its solution is:
\begin{align}
\widetilde{\b{w}}_j^{(t)} = \frac{(\b{G}_j^{(t)})^{-1} \b{1}}{\b{1}^\top (\b{G}_j^{(t)})^{-1} \b{1}},
\end{align}
where $\mathbb{R}^{k \times k} \ni \b{G}_j^{(t)} := (\b{x}_j^{(t)} \b{1}^\top - \b{X}_j^{(t)})^\top (\b{x}_j^{(t)} \b{1}^\top - \b{X}_j^{(t)})$.

The embedding of the out-of-sample $\b{x}_j^{(t)}$ is obtained by the linear representation of the embedding of its $k$ training neighbors:
\begin{align}
\mathbb{R}^p \ni \b{y}_j^{(t)} = \sum_{r=1}^{k} \,_r\widetilde{w}_{j}^{(t)} \,_r\b{y}_j^{(t)},
\end{align}
where $_r\b{y}_j^{(t)}$ is the embedding of the $r$-th training neighbor of $\b{x}_j^{(t)}$.

\section{Supplementary Material: Review of Kernel Locally Linear Embedding}

In the paper, we did not completely review the details of Kernel LLE \cite{zhao2012facial} for the sake of brevity. Here, we review it with a little more details. We do not mention the derivations of equations in LLE because it is out of the scope of this paper.

Kernel LLE \cite{zhao2012facial} finds the $k$-NN graph and performs linear reconstruction from the neighbors in the feature space.
In the paper, we did not completely review the details of kernel LLE \cite{zhao2012facial} for the sake of brevity. Here, we review it with a little more details. We do not mention the derivations of equations in kernel LLE because it is out of the scope of this paper.

\subsection{Embedding The Training Data}

\subsubsection{$k$-Nearest Neighbors}

The Euclidean distance in the feature space is \cite{scholkopf2001kernel}:
\begin{align}
||\b{\phi}(\b{x}_i) - \b{\phi}(\b{x}_j)||_2 &= \sqrt{\big(\b{\phi}(\b{x}_i) - \b{\phi}(\b{x}_j)\big)^\top\big(\b{\phi}(\b{x}_i) - \b{\phi}(\b{x}_j)\big)} \nonumber \\
&= \sqrt{\b{\phi}(\b{x}_i)^\top\b{\phi}(\b{x}_i) -2 \b{\phi}(\b{x}_i)^\top\b{\phi}(\b{x}_j) + \b{\phi}(\b{x}_j)^\top\b{\phi}(\b{x}_j)} \nonumber \\
&= \sqrt{k(\b{x}_i, \b{x}_i) -2 k(\b{x}_i, \b{x}_j) + k(\b{x}_j, \b{x}_j)},
\end{align}
where $\mathbb{R} \ni k(\b{x}_i, \b{x}_j) = \b{\phi}(\b{x}_i)^\top\b{\phi}(\b{x}_j)$ is the $(i,j)$-th element of the kernel matrix $\b{K} \in \mathbb{R}^{n \times n}$.

Using the distances of the data points in the feature space, we construct the $k$-nearest neighbors graph. 
Therefore, every data point has $k$ neighbors in the feature space. Let the matrix $\mathbb{R}^{t \times k} \ni \b{\Phi}(\b{X}_j) := [\b{\phi}(_1\b{x}_{j}), \dots, \b{\phi}(_k\b{x}_{j})]$ include the neighbors of $\b{x}_j$ in the feature space ($t$ is the dimensionality of the feature space).

\subsection{Linear Reconstruction by the Neighbors}

The Eq. (\ref{equation_LLE_linearReconstruct}) in the feature space is:
\begin{equation}\label{equation_kernel_LLE_linearReconstruct}
\begin{aligned}
& \underset{\widetilde{\b{w}}_j}{\text{minimize}}
& & \sum_{j=1}^n \Big|\Big|\b{\phi}(\b{x}_j) - \sum_{r=1}^k \,_r\widetilde{w}_{j}\, \b{\phi}(_r\b{x}_{j})\Big|\Big|_2^2, \\
& \text{subject to}
& & \sum_{r=1}^k \,_r\widetilde{w}_{j} = 1, ~~~ \forall j \in \{1, \dots, n\},
\end{aligned}
\end{equation}
where $\mathbb{R}^k \ni \widetilde{\b{w}}_{j} := [\,_1\widetilde{w}_{j}, \dots, \,_k\widetilde{w}_{j}]^\top$.
The solution to Eq. (\ref{equation_kernel_LLE_linearReconstruct}) is:
\begin{align}
\widetilde{\b{w}}_j = \frac{\b{K}_j^{-1} \b{1}}{\b{1}^\top \b{K}_j^{-1} \b{1}},
\end{align}
where the $(a,b)$-th element of $\b{K}_j \in \mathbb{R}^{k \times k}$ can be calculated as:
\begin{align*}
&\b{K}_j(a,b) = k(\b{x}_j, \b{x}_j) - k(\b{x}_j, \,_a\b{x}_{j}) - k(\b{x}_j, \,_b\b{x}_{j}) + k(_a\b{x}_{j}, \,_b\b{x}_{j}).
\end{align*}

\subsection{Linear Embedding}

The linear embedding in kernel LLE is exactly as the linear embedding in LLE.

\subsection{Embedding The Out-of-sample Data}

The Eq. (\ref{equation_outOfSample_LLE_linearReconstruction}) in the feature space is:
\begin{equation}\label{equation_outOfSample_kernel_LLE_linearReconstruction}
\begin{aligned}
& \underset{\widetilde{\b{w}}_j^{(t)}}{\text{minimize}}
& & \sum_{j=1}^{n_t} \Big|\Big|\b{\phi}(\b{x}_j^{(t)}) - \sum_{r=1}^k \,_r\widetilde{w}_{j}^{(t)} \b{\phi}(_r\b{x}_{j}^{(t)})\Big|\Big|_2^2, \\
& \text{subject to}
& & \sum_{r=1}^k \,_r\widetilde{w}_{j}^{(t)} = 1, ~~~ \forall j \in \{1, \dots, n_t\},
\end{aligned}
\end{equation}
where $\mathbb{R}^k \ni \widetilde{\b{w}}_j^{(t)} := [\,_1\widetilde{w}_{j}^{(t)}, \dots, \,_k\widetilde{w}_{j}^{(t)}]^\top$ includes the weights of linear reconstruction of the $j$-th out-of-sample data point using its $k$ training neighbors in the feature space, and $_r\widetilde{w}_{j}^{(t)}$ is the $r$-th training neighbor of $\b{x}_j$ in the feature space.

The Eq. (\ref{equation_outOfSample_kernel_LLE_linearReconstruction}) is similar to Eq. (\ref{equation_kernel_LLE_linearReconstruct}) and thus its solution is:
\begin{align}
\widetilde{\b{w}}_j^{(t)} = \frac{(\b{K}_j^{(t)})^{-1} \b{1}}{\b{1}^\top (\b{K}_j^{(t)})^{-1} \b{1}},
\end{align}
where the $(a,b)$-th element of $\b{K}_j^{(t)} \in \mathbb{R}^{k \times k}$ can be calculated as:
\begin{align*}
&\b{K}_j^{(t)}(a,b) = k(\b{x}_j^{(t)}, \b{x}_j^{(t)}) - k(\b{x}_j^{(t)}, \,_a\b{x}_{j}^{(t)}) - k(\b{x}_j^{(t)}, \,_b\b{x}_{j}^{(t)}) + k(_a\b{x}_{j}^{(t)}, \,_b\b{x}_{j}^{(t)}).
\end{align*}

The embedding of the out-of-sample $\b{x}_j^{(t)}$ is obtained by the linear representation of the embedding of its $k$ training neighbors:
\begin{align}
\mathbb{R}^p \ni \b{y}_j^{(t)} = \sum_{r=1}^{k} \,_r\widetilde{w}_{j}^{(t)} \,_r\b{y}_j^{(t)},
\end{align}
where $_r\b{y}_j^{(t)}$ is the embedding of the $r$-th training neighbor of $\b{x}_j^{(t)}$ in the feature space.

\section{Supplementary Material: Derivations for Locally Linear Image Structural Embedding}

\subsection{Derivation of Eq. (\ref{equation_LLISE_f})}

In the following, we mention the derivation of Eq. (\ref{equation_LLISE_f}):
\begin{align*}
f(\widetilde{\b{w}}_{j,i}) &= \big|\big|\breve{\b{x}}_{j,i} - \sum_{r=1}^k \,_r\widetilde{w}_{j,i}\, _r\breve{\b{x}}_{j,i}\big|\big|_S = ||\breve{\b{x}}_{j,i} - \breve{\b{X}}_{j,i}\, \widetilde{\b{w}}_{j,i}||_S \\
&\overset{(a)}{=} \frac{||\breve{\b{x}}_{j,i} - \breve{\b{X}}_{j,i}\, \widetilde{\b{w}}_{j,i}||_2^2}{||\breve{\b{x}}_{j,i}||_2^2 + ||\breve{\b{X}}_{j,i}\, \widetilde{\b{w}}_{j,i}||_2^2 + c},
\end{align*}
where $(a)$ is because of Eq. (\ref{equation_SSIM_distance}).
The numerator of $f(\widetilde{\b{w}}_{j,i})$ is simplified as:
\begin{align*}
||\breve{\b{x}}_{j,i} - \breve{\b{X}}_{j,i}\, \widetilde{\b{w}}_{j,i}||_2^2 &=
(\breve{\b{x}}_{j,i} - \breve{\b{X}}_{j,i}\, \widetilde{\b{w}}_{j,i})^\top (\breve{\b{x}}_{j,i} - \breve{\b{X}}_{j,i}\, \widetilde{\b{w}}_{j,i}) \\
&= (\breve{\b{x}}_{j,i}^\top - \widetilde{\b{w}}_{j,i}^\top\, \breve{\b{X}}_{j,i}^\top) (\breve{\b{x}}_{j,i} - \breve{\b{X}}_{j,i}\, \widetilde{\b{w}}_{j,i}) \\
&= \breve{\b{x}}_{j,i}^\top \breve{\b{x}}_{j,i} - \breve{\b{x}}_{j,i}^\top \breve{\b{X}}_{j,i}\, \widetilde{\b{w}}_{j,i} - \widetilde{\b{w}}_{j,i}^\top\, \breve{\b{X}}_{j,i}^\top \breve{\b{x}}_{j,i} + \widetilde{\b{w}}_{j,i}^\top\, \breve{\b{X}}_{j,i}^\top \breve{\b{X}}_{j,i}\, \widetilde{\b{w}}_{j,i} \\
&= \breve{\b{x}}_{j,i}^\top \breve{\b{x}}_{j,i} - \widetilde{\b{w}}_{j,i}^\top\, \breve{\b{X}}_{j,i}^\top \breve{\b{x}}_{j,i} - \widetilde{\b{w}}_{j,i}^\top\, \breve{\b{X}}_{j,i}^\top \breve{\b{x}}_{j,i} + \widetilde{\b{w}}_{j,i}^\top\, \breve{\b{X}}_{j,i}^\top \breve{\b{X}}_{j,i}\, \widetilde{\b{w}}_{j,i} \\
&= \breve{\b{x}}_{j,i}^\top \breve{\b{x}}_{j,i} + \widetilde{\b{w}}_{j,i}^\top\, \breve{\b{X}}_{j,i}^\top \breve{\b{X}}_{j,i}\, \widetilde{\b{w}}_{j,i} -2\, \widetilde{\b{w}}_{j,i}^\top\, \breve{\b{X}}_{j,i}^\top \breve{\b{x}}_{j,i}.
\end{align*}
The first term in denominator of $f(\widetilde{\b{w}}_{j,i})$ is simplified as:
\begin{align*}
||\breve{\b{x}}_{j,i}||_2^2 = \breve{\b{x}}_{j,i}^\top \breve{\b{x}}_{j,i},
\end{align*}
and the second term in denominator of $f(\widetilde{\b{w}}_{j,i})$ is simplified as:
\begin{align*}
||\breve{\b{X}}_{j,i}\, \widetilde{\b{w}}_{j,i}||_2^2 &= (\breve{\b{X}}_{j,i}\, \widetilde{\b{w}}_{j,i})^\top (\breve{\b{X}}_{j,i}\, \widetilde{\b{w}}_{j,i}) =  (\widetilde{\b{w}}_{j,i}^\top\, \breve{\b{X}}_{j,i}^\top) (\breve{\b{X}}_{j,i}\, \widetilde{\b{w}}_{j,i}) \\
&= \widetilde{\b{w}}_{j,i}^\top\, \breve{\b{X}}_{j,i}^\top \breve{\b{X}}_{j,i}\, \widetilde{\b{w}}_{j,i}.
\end{align*}
Therefore, the Eq. (\ref{equation_LLISE_f}) is obtained:
\begin{align}
\mathbb{R} \ni f(\widetilde{\b{w}}_{j,i}) = \frac{\breve{\b{x}}_{j,i}^\top\, \breve{\b{x}}_{j,i} + \widetilde{\b{w}}_{j,i}^\top\, \breve{\b{X}}_{j,i}^\top\, \breve{\b{X}}_{j,i}\, \widetilde{\b{w}}_{j,i} - 2\, \widetilde{\b{w}}_{j,i}^\top\, \breve{\b{X}}_{j,i}^\top\, \breve{\b{x}}_{j,i}}{\breve{\b{x}}_{j,i}^\top\, \breve{\b{x}}_{j,i} + \widetilde{\b{w}}_{j,i}^\top\, \breve{\b{X}}_{j,i}^\top\, \breve{\b{X}}_{j,i}\, \widetilde{\b{w}}_{j,i} + c}.
\end{align}

\subsection{Derivation of Eq. (\ref{equation_LLISE_derivative_f})}

In the following, we mention the derivation of Eq. (\ref{equation_LLISE_derivative_f}).
If we take the numerator and denominator of derivative of $f(\widetilde{\b{w}}_{j,i})$ as:
\begin{align*}
& \mathbb{R} \ni \alpha := \breve{\b{x}}_{j,i}^\top\, \breve{\b{x}}_{j,i} + \widetilde{\b{w}}_{j,i}^\top\, \breve{\b{X}}_{j,i}^\top\, \breve{\b{X}}_{j,i}\, \widetilde{\b{w}}_{j,i} - 2\, \widetilde{\b{w}}_{j,i}^\top\, \breve{\b{X}}_{j,i}^\top\, \breve{\b{x}}_{j,i}, \\
& \mathbb{R} \ni \beta := \breve{\b{x}}_{j,i}^\top\, \breve{\b{x}}_{j,i} + \widetilde{\b{w}}_{j,i}^\top\, \breve{\b{X}}_{j,i}^\top\, \breve{\b{X}}_{j,i}\, \widetilde{\b{w}}_{j,i} + c,
\end{align*}
to have $f(\widetilde{\b{w}}_{j,i}) = \alpha / \beta$, the derivative of $f(\widetilde{\b{w}}_{j,i})$ with respect to $\widetilde{\b{w}}_{j,i}$ is:
\begin{align*}
\mathbb{R}^k \ni \nabla f(\widetilde{\b{w}}_{j,i}) &= \frac{1}{\beta^2}\Big[(\beta)(2\, \breve{\b{X}}_{j,i}^\top\, \breve{\b{X}}_{j,i}\, \widetilde{\b{w}}_{j,i} -2\, \breve{\b{X}}_{j,i}^\top\, \breve{\b{x}}_{j,i}) - (\alpha) (2 \breve{\b{X}}_{j,i}^\top\, \breve{\b{X}}_{j,i}\, \widetilde{\b{w}}_{j,i}) \Big] \\
&= \frac{2}{\beta} (\breve{\b{X}}_{j,i}^\top\, \breve{\b{X}}_{j,i}\, \widetilde{\b{w}}_{j,i} -\, \breve{\b{X}}_{j,i}^\top\, \breve{\b{x}}_{j,i}) - \frac{2\, \alpha}{\beta^2} \breve{\b{X}}_{j,i}^\top\, \breve{\b{X}}_{j,i}\, \widetilde{\b{w}}_{j,i} \\
&= \frac{2}{\beta} \Big(\breve{\b{X}}_{j,i}^\top\, \breve{\b{X}}_{j,i}\, \widetilde{\b{w}}_{j,i} -\, \breve{\b{X}}_{j,i}^\top\, \breve{\b{x}}_{j,i} - f(\widetilde{\b{w}}_{j,i})\, \breve{\b{X}}_{j,i}^\top\, \breve{\b{X}}_{j,i}\, \widetilde{\b{w}}_{j,i}\Big) \\
&= \frac{2}{\beta} \Big(\big(1 - f(\widetilde{\b{w}}_{j,i})\big)\breve{\b{X}}_{j,i}^\top\, \breve{\b{X}}_{j,i}\, \widetilde{\b{w}}_{j,i} -\, \breve{\b{X}}_{j,i}^\top\, \breve{\b{x}}_{j,i}\Big) \\
&= \frac{2\, \breve{\b{X}}_{j,i}^\top}{\beta} \Big(\big(1 - f(\widetilde{\b{w}}_{j,i})\big)\breve{\b{X}}_{j,i}\, \widetilde{\b{w}}_{j,i} -\, \breve{\b{x}}_{j,i}\Big).
\end{align*}
Therefore, the gradient of $f(\widetilde{\b{w}}_{j,i})$ is obtained:
\begin{align}
\mathbb{R}^k \ni \nabla f(\widetilde{\b{w}}_{j,i}) = \frac{2\, \breve{\b{X}}_{j,i}^\top \Big(\big(1 - f(\widetilde{\b{w}}_{j,i})\big) \breve{\b{X}}_{j,i} \widetilde{\b{w}}_{j,i} - \breve{\b{x}}_{j,i}\Big)}{\breve{\b{x}}_{j,i}^\top\, \breve{\b{x}}_{j,i} + \widetilde{\b{w}}_{j,i}^\top\, \breve{\b{X}}_{j,i}^\top\, \breve{\b{X}}_{j,i}\, \widetilde{\b{w}}_{j,i} + c}.
\end{align}

\subsection{Derivation of Update of $\widetilde{\b{w}}_{j,i}$ in Eq. (\ref{equation_LLISE_reconst_ADMM_updates})}

In the following, we mention the derivation of Eq. (\ref{equation_LLISE_reconst_ADMM_updates}).
The Eq. (\ref{equation_LLISE_reconst_ADMM_w_update}) is:
\begin{align*}
\widetilde{\b{w}}_{j,i}^{(\nu+1)} & := \arg \min_{\widetilde{\b{w}}_{j,i}} \Big( f(\widetilde{\b{w}}_{j,i}) + (\rho/2)\, ||\widetilde{\b{w}}_{j,i} - \widetilde{\b{\xi}}_{j,i}^{(\nu)} + \b{j}_{j,i}^{(\nu)}||_2^2 \Big).
\end{align*}
The objective function can be simplified as:
\begin{align*}
f(\widetilde{\b{w}}_{j,i}) &+ (\rho/2)\, ||\widetilde{\b{w}}_{j,i} - \widetilde{\b{\xi}}_{j,i}^{(\nu)} + \b{j}_{j,i}^{(\nu)}||_2^2 \\
&= f(\widetilde{\b{w}}_{j,i}) + (\rho/2)\, \Big( (\widetilde{\b{w}}_{j,i} - \widetilde{\b{\xi}}_{j,i}^{(\nu)} + \b{j}_{j,i}^{(\nu)})^\top (\widetilde{\b{w}}_{j,i} - \widetilde{\b{\xi}}_{j,i}^{(\nu)} + \b{j}_{j,i}^{(\nu)}) \Big) \\
&= f(\widetilde{\b{w}}_{j,i}) + (\rho/2)\, \Big( (\widetilde{\b{w}}_{j,i}^\top - \widetilde{\b{\xi}}_{j,i}^{(\nu)\top} + \b{j}_{j,i}^{(\nu)\top}) (\widetilde{\b{w}}_{j,i} - \widetilde{\b{\xi}}_{j,i}^{(\nu)} + \b{j}_{j,i}^{(\nu)}) \Big) \\
&= f(\widetilde{\b{w}}_{j,i}) + (\rho/2)\, \Big( \widetilde{\b{w}}_{j,i}^\top\, \widetilde{\b{w}}_{j,i} - \widetilde{\b{w}}_{j,i}^\top\, \widetilde{\b{\xi}}_{j,i}^{(\nu)} + \widetilde{\b{w}}_{j,i}^\top\, \b{j}_{j,i}^{(\nu)} - \widetilde{\b{\xi}}_{j,i}^{(\nu)\top} \widetilde{\b{w}}_{j,i} \\
&+ \widetilde{\b{\xi}}_{j,i}^{(\nu)\top} \widetilde{\b{\xi}}_{j,i}^{(\nu)} - \widetilde{\b{\xi}}_{j,i}^{(\nu)\top} \b{j}_{j,i}^{(\nu)} + \b{j}_{j,i}^{(\nu)\top} \widetilde{\b{w}}_{j,i} - \b{j}_{j,i}^{(\nu)\top} \widetilde{\b{\xi}}_{j,i}^{(\nu)} + \b{j}_{j,i}^{(\nu)\top} \b{j}_{j,i}^{(\nu)} \Big).
\end{align*}
The gradient of the objective function with respect to $\b{U}_i$ is:
\begin{align*}
\frac{\partial}{\partial \widetilde{\b{w}}_{j,i}}\Big(& f(\widetilde{\b{w}}_{j,i}) + (\rho/2)\, ||\widetilde{\b{w}}_{j,i} - \widetilde{\b{\xi}}_{j,i}^{(\nu)} + \b{j}_{j,i}^{(\nu)}||_2^2 \Big) \\
&= \nabla f(\widetilde{\b{w}}_{j,i}) + (\rho/2)\, \Big( 2\,\widetilde{\b{w}}_{j,i} - \widetilde{\b{\xi}}_{j,i}^{(\nu)} + \b{j}_{j,i}^{(\nu)} - \widetilde{\b{\xi}}_{j,i}^{(\nu)} + \b{j}_{j,i}^{(\nu)} \Big) \\
&= \nabla f(\widetilde{\b{w}}_{j,i}) + \rho\, ( \widetilde{\b{w}}_{j,i} - \widetilde{\b{\xi}}_{j,i}^{(\nu)} + \b{j}_{j,i}^{(\nu)} ).
\end{align*}
Therefore, the iteration in gradient descent is:
\begin{align}
\widetilde{\b{w}}_{j,i}^{(\nu+1)} & := \widetilde{\b{w}}_{j,i}^{(\nu)} - \eta\, \frac{\partial}{\partial \widetilde{\b{w}}_{j,i}}( ... ) = \widetilde{\b{w}}_{j,i}^{(\nu)} - \eta\, \nabla f(\widetilde{\b{w}}_{j,i}^{(\nu)}) - \eta\,\rho\, (\widetilde{\b{w}}_{j,i}^{(\nu)} - \widetilde{\b{\xi}}_{j,i}^{(\nu)} + \b{j}_{j,i}^{(\nu)}),
\end{align}
where $\eta$ is the learning rate and $\frac{\partial}{\partial \widetilde{\b{w}}_{j,i}}( ... )$ is derivative of the objective function.

\subsection{Derivation of Eq. (\ref{equation_LLISE_theta_Y})}

In the following, we mention the derivation of Eq. (\ref{equation_LLISE_theta_Y}):
\begin{align*}
\theta_j(\b{\b{Y}}_i) := ||\b{Y}_i^\top \b{1}_j - \b{Y}_i^\top \b{w}_{j,i}||_S \overset{(a)}{=} \frac{||\b{Y}_i^\top \b{1}_j - \b{Y}_i^\top \b{w}_{j,i}||_2^2}{||\b{Y}_i^\top \b{1}_j||_2^2 + ||\b{Y}_i^\top \b{w}_{j,i}||_2^2 + c},
\end{align*}
where $(a)$ is because of Eq. (\ref{equation_SSIM_distance}).
The numerator of $\theta_j(\b{\b{Y}}_i)$ is simplified as:
\begin{align*}
||\b{Y}_i^\top \b{1}_j - &\b{Y}_i^\top \b{w}_{j,i}||_2^2 = (\b{Y}_i^\top \b{1}_j - \b{Y}_i^\top \b{w}_{j,i})^\top (\b{Y}_i^\top \b{1}_j - \b{Y}_i^\top \b{w}_{j,i}) \\
&= (\b{1}_j^\top \b{Y}_i  - \b{w}_{j,i}^\top \b{Y}_i) (\b{Y}_i^\top \b{1}_j - \b{Y}_i^\top \b{w}_{j,i}) \\
&= \b{1}_j^\top \b{Y}_i \b{Y}_i^\top \b{1}_j - \b{1}_j^\top \b{Y}_i \b{Y}_i^\top \b{w}_{j,i} - \b{w}_{j,i}^\top \b{Y}_i \b{Y}_i^\top \b{1}_j + \b{w}_{j,i}^\top \b{Y}_i \b{Y}_i^\top \b{w}_{j,i} \\
&= \b{1}_j^\top \b{Y}_i \b{Y}_i^\top \b{1}_j - \b{w}_{j,i}^\top \b{Y}_i \b{Y}_i^\top \b{1}_j - \b{w}_{j,i}^\top \b{Y}_i \b{Y}_i^\top \b{1}_j + \b{w}_{j,i}^\top \b{Y}_i \b{Y}_i^\top \b{w}_{j,i} \\
&= \b{1}_j^\top \b{Y}_i \b{Y}_i^\top \b{1}_j + \b{w}_{j,i}^\top \b{Y}_i \b{Y}_i^\top \b{w}_{j,i} -2\, \b{w}_{j,i}^\top \b{Y}_i \b{Y}_i^\top \b{1}_j.
\end{align*}
The numerator of $\theta_j(\b{\b{Y}}_i)$ is a scalar so it is equal to its trace (we denote trace of matrix by $\textbf{tr}(.)$):
\begin{align*}
||\b{Y}_i^\top \b{1}_j - &\b{Y}_i^\top \b{w}_{j,i}||_2^2 = \textbf{tr}(\b{1}_j^\top \b{Y}_i \b{Y}_i^\top \b{1}_j + \b{w}_{j,i}^\top \b{Y}_i \b{Y}_i^\top \b{w}_{j,i} -2\, \b{w}_{j,i}^\top \b{Y}_i \b{Y}_i^\top \b{1}_j) \\
&= \textbf{tr}(\b{1}_j^\top \b{Y}_i \b{Y}_i^\top \b{1}_j) + \textbf{tr}(\b{w}_{j,i}^\top \b{Y}_i \b{Y}_i^\top \b{w}_{j,i}) -2\, \textbf{tr}(\b{w}_{j,i}^\top \b{Y}_i \b{Y}_i^\top \b{1}_j) \\
&\overset{(a)}{=} \textbf{tr}(\b{Y}_i^\top \b{1}_j\, \b{1}_j^\top \b{Y}_i) + \textbf{tr}(\b{Y}_i^\top \b{w}_{j,i}\, \b{w}_{j,i}^\top \b{Y}_i) -2\, \textbf{tr}(\b{Y}_i^\top \b{1}_j\, \b{w}_{j,i}^\top \b{Y}_i) \\
&= \textbf{tr}(\b{Y}_i^\top \b{1}_j\, \b{1}_j^\top \b{Y}_i + \b{Y}_i^\top \b{w}_{j,i}\, \b{w}_{j,i}^\top \b{Y}_i -2\, \b{Y}_i^\top \b{1}_j\, \b{w}_{j,i}^\top \b{Y}_i) \\
&= \textbf{tr}\big(\b{Y}_i^\top (\b{1}_j\, \b{1}_j^\top + \b{w}_{j,i}\, \b{w}_{j,i}^\top -2\, \b{1}_j\, \b{w}_{j,i}^\top) \b{Y}_i\big) = \textbf{tr}(\b{Y}_i^\top \b{M}_{j,i}\, \b{Y}_i),
\end{align*}
where $(a)$ is because of the cyclic property of trace and $\mathbb{R}^{n \times n} \ni \b{M}_{j,i} := \b{1}_j \b{1}_j^\top + \b{w}_{j,i} \b{w}_{j,i}^\top - 2\, \b{1}_j \b{w}_{j,i}^\top$.

The first term in denominator of $\theta_j(\b{\b{Y}}_i)$ is simplified as:
\begin{align*}
||\b{Y}_i^\top \b{1}_j||_2^2 = (\b{Y}_i^\top \b{1}_j)^\top (\b{Y}_i^\top \b{1}_j) = (\b{1}_j^\top \b{Y}_i) (\b{Y}_i^\top \b{1}_j) = \b{1}_j^\top \b{Y}_i \b{Y}_i^\top \b{1}_j,
\end{align*}
and the second term in denominator of $\theta_j(\b{\b{Y}}_i)$ is simplified as:
\begin{align*}
||\b{Y}_i^\top \b{w}_{j,i}||_2^2 = (\b{Y}_i^\top \b{w}_{j,i})^\top (\b{Y}_i^\top \b{w}_{j,i}) = (\b{w}_{j,i}^\top \b{Y}_i) (\b{Y}_i^\top \b{w}_{j,i}) = \b{w}_{j,i}^\top \b{Y}_i \b{Y}_i^\top \b{w}_{j,i}.
\end{align*}
Therefore, the denominator of $\theta_j(\b{\b{Y}}_i)$ is:
\begin{align*}
||\b{Y}_i^\top \b{1}_j||_2^2 + ||\b{Y}_i^\top \b{w}_{j,i}||_2^2 + c = \b{1}_j^\top \b{Y}_i \b{Y}_i^\top \b{1}_j + \b{w}_{j,i}^\top \b{Y}_i \b{Y}_i^\top \b{w}_{j,i} + c,
\end{align*}
which is a scalar so it is equal to its trace:
\begin{align*}
||\b{Y}_i^\top & \b{1}_j||_2^2 + ||\b{Y}_i^\top \b{w}_{j,i}||_2^2 + c = \textbf{tr}(\b{1}_j^\top \b{Y}_i \b{Y}_i^\top \b{1}_j + \b{w}_{j,i}^\top \b{Y}_i \b{Y}_i^\top \b{w}_{j,i} + c) \\
&= \textbf{tr}(\b{1}_j^\top \b{Y}_i \b{Y}_i^\top \b{1}_j) + \textbf{tr}(\b{w}_{j,i}^\top \b{Y}_i \b{Y}_i^\top \b{w}_{j,i}) + c \\
&= \textbf{tr}(\b{Y}_i^\top \b{1}_j\, \b{1}_j^\top \b{Y}_i) + \textbf{tr}(\b{Y}_i^\top \b{w}_{j,i}\, \b{w}_{j,i}^\top \b{Y}_i) + c \\
&= \textbf{tr}(\b{Y}_i^\top \b{1}_j\, \b{1}_j^\top \b{Y}_i + \b{Y}_i^\top \b{w}_{j,i}\, \b{w}_{j,i}^\top \b{Y}_i + c) \\
&= \textbf{tr}\big(\b{Y}_i^\top (\b{1}_j\, \b{1}_j^\top + \b{w}_{j,i}\, \b{w}_{j,i}^\top) \b{Y}_i + c\big) \\
&= \textbf{tr}(\b{Y}_i^\top \b{1}_j\, \b{1}_j^\top \b{Y}_i + \b{Y}_i^\top \b{w}_{j,i}\, \b{w}_{j,i}^\top \b{Y}_i + c) = \textbf{tr}(\b{Y}_i^\top \b{\Psi}_{j,i}\, \b{Y}_i + c),
\end{align*}
where $\mathbb{R}^{n \times n} \ni \b{\Psi}_{j,i} := \b{1}_j \b{1}_j^\top + \b{w}_{j,i} \b{w}_{j,i}^\top = \b{M}_{j,i} + 2\, \b{1}_j \b{w}_{j,i}^\top$.

Therefore, the Eq. (\ref{equation_LLISE_theta_Y}) is obtained:
\begin{align}
\mathbb{R} \ni \theta_j(\b{Y}_i) = \frac{\textbf{tr}(\b{Y}_i^\top \b{M}_{j,i}\, \b{Y}_i)}{\textbf{tr}(\b{Y}_i^\top \b{\Psi}_{j,i}\, \b{Y}_i) + c}.
\end{align}

\subsection{Derivation of Eq. (\ref{equation_LLISE_gradient_theta_Y})}

In the following, we mention the derivation of Eq. (\ref{equation_LLISE_gradient_theta_Y}).
If we take the numerator and denominator of derivative of $\theta_j(\b{Y}_i)$ as:
\begin{align*}
& \mathbb{R} \ni \alpha := \textbf{tr}(\b{Y}_i^\top \b{M}_{j,i}\, \b{Y}_i), \\
& \mathbb{R} \ni \beta := \textbf{tr}(\b{Y}_i^\top \b{\Psi}_{j,i}\, \b{Y}_i) + c,
\end{align*}
to have $\theta_j(\b{Y}_i) = \alpha / \beta$, the derivative of $\theta_j(\b{Y}_i)$ with respect to $\b{Y}_i$ is:
\begin{align*}
\mathbb{R}^{n \times p} \ni \nabla \theta_j(\b{Y}_i) &= \frac{1}{\beta^2}\Big[(\beta)(2\, \b{M}_{j,i}\, \b{Y}_i) - (\alpha) (2\, \b{\Psi}_{j,i}\, \b{Y}_i) \Big] \\
&= \frac{2\, \b{M}_{j,i}\, \b{Y}_i}{\beta} - \frac{2\,\alpha}{\beta^2} (\b{\Psi}_{j,i}\, \b{Y}_i) \\
&= \frac{2}{\beta} \Big[\b{M}_{j,i}\, \b{Y}_i - \theta_j(\b{Y}_i)\, \b{\Psi}_{j,i}\, \b{Y}_i\Big] = \frac{2}{\beta} \Big[\b{M}_{j,i} - \theta_j(\b{Y}_i)\, \b{\Psi}_{j,i}\Big] \b{Y}_i
\end{align*}
Therefore, the gradient of $\theta_j(\b{Y}_i)$ is obtained:
\begin{align}
\mathbb{R}^{n \times p} \ni \nabla \theta_j(\b{Y}_i) = \frac{2}{\textbf{tr}(\b{Y}_i^\top \b{\Psi}_{j,i}\, \b{Y}_i) + c} \Big(\b{M}_{j,i} - \theta_j(\b{Y}_i)\,\b{\Psi}_{j,i}\Big)\, \b{Y}_i.
\end{align}

\subsection{Derivation of Update of $\b{Y}_i$ in Eq. (\ref{equation_LLISE_ADMM_updates})}

In the following, we mention the derivation of Eq. (\ref{equation_LLISE_ADMM_updates}).
The Eq. (\ref{equation_LLISE_ADMM_Y_update}) is:
\begin{align*}
\b{Y}_i^{(\nu+1)} & := \arg \min_{\b{Y}_i} \Big( \sum_{j=1}^n \big( \theta_j(\b{Y}_i) \big) + (\rho/2)\, ||\b{Y}_i - \b{V}_i^{(\nu)} + \b{J}_i^{(\nu)}||_F^2 \Big).
\end{align*}
The objective function can be simplified as:
\begin{align*}
&\sum_{j=1}^n \big( \theta_j(\b{Y}_i) \big) + (\rho/2)\, ||\b{Y}_i - \b{V}_i^{(\nu)} + \b{J}_i^{(\nu)}||_F^2 \\
&= \sum_{j=1}^n \big( \theta_j(\b{Y}_i) \big) + (\rho/2)\, \textbf{tr} \Big( (\b{Y}_i - \b{V}_i^{(\nu)} + \b{J}_i^{(\nu)})^\top (\b{Y}_i - \b{V}_i^{(\nu)} + \b{J}_i^{(\nu)}) \Big) \\
&= \sum_{j=1}^n \big( \theta_j(\b{Y}_i) \big) + (\rho/2)\, \textbf{tr} \Big( (\b{Y}_i^\top - \b{V}_i^{(\nu)\top} + \b{J}_i^{(\nu)\top}) (\b{Y}_i - \b{V}_i^{(\nu)} + \b{J}_i^{(\nu)}) \Big) \\
&= \sum_{j=1}^n \big( \theta_j(\b{Y}_i) \big) + (\rho/2)\, \textbf{tr} \Big( \b{Y}_i^\top \b{Y}_i - \b{Y}_i^\top \b{V}_i^{(\nu)} + \b{Y}_i^\top \b{J}_i^{(\nu)} - \b{V}_i^{(\nu)\top} \b{Y}_i \\
&+ \b{V}_i^{(\nu)\top}\b{V}_i^{(\nu)} - \b{V}_i^{(\nu)\top} \b{J}_i^{(\nu)} + \b{J}_i^{(\nu)\top} \b{Y}_i - \b{J}_i^{(\nu)\top} \b{V}_i^{(\nu)} + \b{J}_i^{(\nu)\top} \b{J}_i^{(\nu)} \Big).
\end{align*}
The gradient of the objective function with respect to $\b{Y}_i$ is:
\begin{align*}
&\frac{\partial}{\partial \b{Y}_i}\Big( \sum_{j=1}^n \big( \theta_j(\b{Y}_i) \big) + (\rho/2)\, ||\b{Y}_i - \b{V}_i^{(\nu)} + \b{J}_i^{(\nu)}||_F^2 \Big) \\
&= \sum_{j=1}^n \big(\nabla \theta_j(\b{Y}_i)\big) + (\rho/2)\, ( 2\,\b{Y}_i - \b{V}_i^{(\nu)} + \b{J}_i^{(\nu)} - \b{V}_i^{(\nu)}  + \b{J}_i^{(\nu)} ) \\
&= \sum_{j=1}^n \big(\nabla \theta_j(\b{Y}_i)\big) + \rho\, ( \b{Y}_i - \b{V}_i^{(\nu)} + \b{J}_i^{(\nu)} ).
\end{align*}
Therefore, the iteration in gradient descent is:
\begin{align}
\b{Y}_i^{(\nu+1)} & := \b{Y}_i^{(\nu)} - \eta\, \frac{\partial}{\partial \b{Y}_i}( ... ) \nonumber \\
&= \b{Y}_i^{(\nu)} - \eta\, \sum_{j=1}^n \big(\nabla \theta_j(\b{Y}_i)\big) - \eta\,\rho\, (\b{Y}_i - \b{V}_i^{(\nu)} + \b{J}_i^{(\nu)}),
\end{align}
where $\eta$ is the learning rate and $\frac{\partial}{\partial \b{Y}_i}( ... )$ is derivative of the objective function.

\section{Supplementary Material: Derivations for Kernel Locally Linear Image Structural Embedding}

\subsection{Derivation of Eq. (\ref{equation_kernel_LLISE_f})}

In the following, we mention the derivation of Eq. (\ref{equation_kernel_LLISE_f}):
\begin{align*}
f^{\phi}(\b{\widetilde{w}}_{j,i}) &= \big|\big|\b{\phi}(\breve{\b{x}}_{j,i}) - \sum_{r=1}^k \,_r\widetilde{w}_{ij} \, \b{\phi}(_r\breve{\b{x}}_{j,i})\big|\big|_S = \big|\big|\b{\phi}(\breve{\b{x}}_{j,i}) - \b{\Phi}(\breve{\b{X}}_{j,i})\, \b{\widetilde{w}}_{j,i}\big|\big|_S \\
&\overset{(a)}{=} \frac{||\b{\phi}(\breve{\b{x}}_{j,i}) - \b{\Phi}(\breve{\b{X}}_{j,i})\, \b{\widetilde{w}}_{j,i}||_2^2}{||\b{\phi}(\breve{\b{x}}_{j,i})||_2^2 + ||\b{\Phi}(\breve{\b{X}}_{j,i})\, \b{\widetilde{w}}_{j,i}||_2^2 + c},
\end{align*}
where $(a)$ is because of Eq. (\ref{equation_SSIM_distance}).
The numerator of $f^{\phi}(\b{\widetilde{w}}_{j,i})$ is simplified as:
\begin{align*}
&||\b{\phi}(\breve{\b{x}}_{j,i}) - \b{\Phi}(\breve{\b{X}}_{j,i})\, \b{\widetilde{w}}_{j,i}||_2^2 \\
&= \big(\b{\phi}(\breve{\b{x}}_{j,i}) - \b{\Phi}(\breve{\b{X}}_{j,i})\, \b{\widetilde{w}}_{j,i}\big)^\top \big(\b{\phi}(\breve{\b{x}}_{j,i}) - \b{\Phi}(\breve{\b{X}}_{j,i})\, \b{\widetilde{w}}_{j,i}\big) \\
&= \big(\b{\phi}(\breve{\b{x}}_{j,i})^\top - \b{\widetilde{w}}_{j,i}^\top\,\b{\Phi}(\breve{\b{X}}_{j,i})^\top\big) \big(\b{\phi}(\breve{\b{x}}_{j,i}) - \b{\Phi}(\breve{\b{X}}_{j,i})\, \b{\widetilde{w}}_{j,i}\big) \\
&= \b{\phi}(\breve{\b{x}}_{j,i})^\top \b{\phi}(\breve{\b{x}}_{j,i}) - \b{\phi}(\breve{\b{x}}_{j,i})^\top \b{\Phi}(\breve{\b{X}}_{j,i})\, \b{\widetilde{w}}_{j,i} - \b{\widetilde{w}}_{j,i}^\top\,\b{\Phi}(\breve{\b{X}}_{j,i})^\top \b{\phi}(\breve{\b{x}}_{j,i}) \\
&~~~~~~~~~~~~~~~~~~~~~~~~~~~~~~~~~~~~~~~~~~~~~~~~~~~~~~~~ + \b{\widetilde{w}}_{j,i}^\top\,\b{\Phi}(\breve{\b{X}}_{j,i})^\top \b{\Phi}(\breve{\b{X}}_{j,i})\, \b{\widetilde{w}}_{j,i} \\
&= \b{\phi}(\breve{\b{x}}_{j,i})^\top \b{\phi}(\breve{\b{x}}_{j,i}) - \b{\widetilde{w}}_{j,i}^\top\,\b{\Phi}(\breve{\b{X}}_{j,i})^\top \b{\phi}(\breve{\b{x}}_{j,i}) - \b{\widetilde{w}}_{j,i}^\top\,\b{\Phi}(\breve{\b{X}}_{j,i})^\top \b{\phi}(\breve{\b{x}}_{j,i}) \\
&~~~~~~~~~~~~~~~~~~~~~~~~~~~~~~~~~~~~~~~~~~~~~~~~~~~~~~~~ + \b{\widetilde{w}}_{j,i}^\top\,\b{\Phi}(\breve{\b{X}}_{j,i})^\top \b{\Phi}(\breve{\b{X}}_{j,i})\, \b{\widetilde{w}}_{j,i} \\
&= k_{j,i} + \widetilde{\b{w}}_{j,i}^\top\, \b{K}_{j,i}\, \widetilde{\b{w}}_{j,i} - 2\, \widetilde{\b{w}}_{j,i}^\top\, \b{k}_{j,i}.
\end{align*}
The first term in denominator of $f^{\phi}(\b{\widetilde{w}}_{j,i})$ is simplified as:
\begin{align*}
||\b{\phi}(\breve{\b{x}}_{j,i})||_2^2 = \b{\phi}(\breve{\b{x}}_{j,i})^\top \b{\phi}(\breve{\b{x}}_{j,i}) = k_{j,i},
\end{align*}
and the second term in denominator of $f^{\phi}(\b{\widetilde{w}}_{j,i})$ is simplified as:
\begin{align*}
||\b{\Phi}(\breve{\b{X}}_{j,i})\, \b{\widetilde{w}}_{j,i}||_2^2 &= \big(\b{\Phi}(\breve{\b{X}}_{j,i})\, \b{\widetilde{w}}_{j,i}\big)^\top \big(\b{\Phi}(\breve{\b{X}}_{j,i})\, \b{\widetilde{w}}_{j,i}\big) \\
&= \big(\b{\widetilde{w}}_{j,i}^\top\, \b{\Phi}(\breve{\b{X}}_{j,i})^\top\big) \big(\b{\Phi}(\breve{\b{X}}_{j,i})\, \b{\widetilde{w}}_{j,i}\big) \\
&= \b{\widetilde{w}}_{j,i}^\top\, \b{\Phi}(\breve{\b{X}}_{j,i})^\top \b{\Phi}(\breve{\b{X}}_{j,i})\, \b{\widetilde{w}}_{j,i} = \widetilde{\b{w}}_{j,i}^\top\, \b{K}_{j,i}\, \widetilde{\b{w}}_{j,i}.
\end{align*}
Therefore, the Eq. (\ref{equation_kernel_LLISE_f}) is obtained:
\begin{align}
\mathbb{R} \ni f^{\phi}(\widetilde{\b{w}}_{j,i}) = \frac{k_{j,i} + \widetilde{\b{w}}_{j,i}^\top\, \b{K}_{j,i}\, \widetilde{\b{w}}_{j,i} - 2\, \widetilde{\b{w}}_{j,i}^\top\, \b{k}_{j,i}}{k_{j,i} + \widetilde{\b{w}}_{j,i}^\top\, \b{K}_{j,i}\, \widetilde{\b{w}}_{j,i} + c}.
\end{align}

\subsection{Derivation of Eq. (\ref{equation_kernel_LLISE_derivative_f})}

In the following, we mention the derivation of Eq. (\ref{equation_kernel_LLISE_derivative_f}).
If we take the numerator and denominator of derivative of $f^{\phi}(\widetilde{\b{w}}_{j,i})$ as:
\begin{align*}
& \mathbb{R} \ni \alpha := k_{j,i} + \widetilde{\b{w}}_{j,i}^\top\, \b{K}_{j,i}\, \widetilde{\b{w}}_{j,i} - 2\, \widetilde{\b{w}}_{j,i}^\top\, \b{k}_{j,i}, \\
& \mathbb{R} \ni \beta := k_{j,i} + \widetilde{\b{w}}_{j,i}^\top\, \b{K}_{j,i}\, \widetilde{\b{w}}_{j,i} + c,
\end{align*}
to have $f^{\phi}(\widetilde{\b{w}}_{j,i}) = \alpha / \beta$, the derivative of $f^{\phi}(\widetilde{\b{w}}_{j,i})$ with respect to $\widetilde{\b{w}}_{j,i}$ is:
\begin{align*}
\mathbb{R}^k \ni \nabla f^{\phi}(\widetilde{\b{w}}_{j,i}) &= \frac{1}{\beta^2}\Big[(\beta)(2\, \b{K}_{j,i}\, \widetilde{\b{w}}_{j,i} - 2\, \b{k}_{j,i}) - (\alpha) (2\, \b{K}_{j,i}\, \widetilde{\b{w}}_{j,i}) \Big] = \\
&= \frac{2}{\beta^2}\Big[(\beta - \alpha)(\b{K}_{j,i}\, \widetilde{\b{w}}_{j,i}) - \beta\, \b{k}_{j,i} \Big] = \big(\frac{2}{\beta} - \frac{2}{\beta} f^{\phi}(\widetilde{\b{w}}_{j,i})\big)(\b{K}_{j,i}\, \widetilde{\b{w}}_{j,i}) - \frac{2}{\beta}\, \b{k}_{j,i} \\
&= \frac{2}{\beta} \Big[ \big(1 - f^{\phi}(\widetilde{\b{w}}_{j,i})\big)\b{K}_{j,i}\, \widetilde{\b{w}}_{j,i} - \b{k}_{j,i}\Big].
\end{align*}
Therefore, the gradient of $f^{\phi}(\widetilde{\b{w}}_{j,i})$ is obtained:
\begin{align}
\mathbb{R}^k \ni \nabla f^{\phi}(\widetilde{\b{w}}_{j,i}) = \frac{2\, \Big(\big(1 - f^{\phi}(\widetilde{\b{w}}_{j,i})\big) \b{K}_{j,i}\, \widetilde{\b{w}}_{j,i} - \b{k}_{j,i}\Big)}{k_{j,i} + \widetilde{\b{w}}_{j,i}^\top\, \b{K}_{j,i}\, \widetilde{\b{w}}_{j,i} + c}.
\end{align}

\end{document}